\documentclass[journal]{IEEEtran}
\ifCLASSINFOpdf
\else
\fi
\usepackage{amssymb}
\usepackage{tabularx}
\usepackage{amstext}
\usepackage{amsmath}
\usepackage{amsthm}
\usepackage{epsfig}
\usepackage{epstopdf}
\usepackage{calc}
\usepackage{color}
\usepackage{graphicx}
\usepackage{caption}
\usepackage{booktabs}
\usepackage{rotating}
\usepackage{pdflscape}
\usepackage{slashbox}
\usepackage{subfig}
\usepackage{array}

\newcommand{\sgn}{\operatorname{sgn}}

\begin{document}
\title{Stochastic Gradient Based Extreme Learning Machines For Online Learning of Advanced Combustion Engines}
\author{Vijay Manikandan Janakiraman,
        XuanLong Nguyen,
        and Dennis Assanis 
\thanks{Vijay Manikandan Janakiraman is currently with UARC @ NASA Ames Research Center, Moffett Field, CA, USA. He was previously with the Department of Mechanical Engineering, University of Michigan, Ann Arbor, MI, USA. E-mail: vijai@umich.edu.}
\thanks{XuanLong Nguyen is with the Department of Statistics, University of Michigan, Ann Arbor, MI, USA.}
\thanks{Dennis Assanis is with the Stony Brook University, NY, USA.}}

\markboth{IEEE TRANSACTIONS ON NEURAL NETWORKS AND LEARNING SYSTEMS, VOL. XX, NO. XX, XXXX 2014}%
{Janakiraman \MakeLowercase{\textit{et al.}}: Stochastic Gradient Based Extreme Learning Machines For Online Learning of Advanced Combustion Engines}
\maketitle

\begin{abstract}
In this article, a stochastic gradient based online learning algorithm for Extreme Learning Machines (ELM) is developed (SG-ELM). A stability criterion based on Lyapunov approach is used to prove both asymptotic stability of estimation error and stability in the estimated parameters suitable for identification of nonlinear dynamic systems. The developed algorithm not only guarantees stability, but also reduces the computational demand compared to the OS-ELM approach \cite{oselm} based on recursive least squares. In order to demonstrate the effectiveness of the algorithm on a real-world scenario, an advanced combustion engine identification problem is considered. The algorithm is applied to two case studies: An online regression learning for system identification of a Homogeneous Charge Compression Ignition (HCCI) Engine and an online classification learning (with class imbalance) for identifying the dynamic operating envelope of the HCCI Engine. The results indicate that the accuracy of the proposed SG-ELM is comparable to that of the state-of-the-art but adds stability and a reduction in computational effort.
\end{abstract}


\section{Introduction} \label{intro}
Homogeneous Charge Compression Ignition (HCCI) Engines are of significant interest to the automotive industry owing to their ability to reduce emissions and fuel consumption significantly compared to traditional spark ignition and compression ignition engines \cite{thring1,Christensen2,Aoyama3}. The highly efficient operation of HCCI is achieved using advanced control strategies such as exhaust gas recirculation (EGR) \cite{kyoungjoon}, variable valve timings (VVT) \cite{Johansson2010}, intake charge heating \cite{heatedintake} among others. Such complex manipulations of the system results in a highly nonlinear behavior \cite{nonlin_HCCI} with a narrow region of stable operation \cite{narrow1,narrow2}.

Control of HCCI combustion is a major challenge for automotive application. Several factors contribute to the challenge including the absence of a direct trigger for combustion, narrow operating range and high sensitivity to disturbances. To address the issue, advanced model based control methods are common where the control actions are often made using a predictive model of the engine \cite{chiang2010,Johansson2010,Ravi2009}. As alternatives to physics based modeling that might involve significant development time and associated costs, data based approaches were introduced \cite{vijay_asoc,vijay_springer,vijay_open} that takes advantage of the extensive experimentation that is performed during the engine calibration process.

The key requirement for a model based control of an HCCI engine is the ability to accurately predict the engine state variables for several operating cycles ahead of time, so that a control action with a known effect can be applied to the engine. Further, in order to be vigilant against the engine drifting towards instabilities such as misfire, ringing, knock, etc \cite{misf1,misf2}, the operating limits of the engine particularly in transients, is required. In order to develop controllers and operate the engine in a stable manner, both models of the engine operating envelope as well as models of engine state variables are necessary.

The state variables of an engine can be defined as the fundamental quantities that represent the state of operation of the engine. As a consequence, these variables also influence the performance of the engine such as fuel efficiency, emissions and stability, and are required to be monitored/regulated. For this work, the net mean effective pressure (NMEP) and the phasing of combustion event (CA50) with respect to the engine's top dead center \cite{vijay_asoc} are considered representative states that represents the quality of engine operation. More fundamental state variables such as in-cylinder temperature, pressure, chemical composition of combustion mixtures can be considered but these variables cannot be measured feasibly on a production engine.

The HCCI engine has a narrow region of stable operation defined by an operating envelope. The dynamic operating envelope of an engine can be defined as a stable region in the operating space of the engine. The significance of the operating envelope and data based modeling approaches are recently introduced by the authors \cite{vijay_open}. Knowledge of the operating envelope is crucial for designing efficient controllers for the following reasons. The developer can get insights on the actuator extremes \cite{svm_ilya}, such as the minimum and maximum quantity of fuel to be injected into the engine at a given speed and load conditions. The actuator extremes can then be used to enforce constraints on the control variables for desired engine operation. Furthermore, an operating envelope model could enable designing efficient engine diagnostic systems based on predictive analytics. For instance, a misfire event is a lack of combustion which produces no work output from the engine. The misfired fuel enters the exhaust system increasing emissions of hydrocarbon and carbon monoxide \cite{misfire_emissions,misfire_emission}. When the engine misfires, pollutant levels may be higher than normal. Real time monitoring of the exhaust emission control system and engine misfire detection are essential to meet requirements on On-Board Diagnostic (OBD) regulations. The envelope model can be used to alarm the onboard diagnostics if the engine is about to misfire owing to changes in system or operating conditions.

Data based modeling approaches for the HCCI engine state variables and dynamic operating envelope were demonstrated using neural networks \cite{vijay_asoc}, support vector machines \cite{vijay_springer}, extreme learning machines \cite{vijay_control} by the authors. However, the previous research considered an offline approach where the data collected from engine experiments were taken offline and models were developed using computer workstations that had high processing and memory. However, a key requirement in advancing the capabilities of data based HCCI modeling task is to perform online learning for the following reasons. The models developed offline are valid only in the controlled experimental conditions. For instance, the experiments are performed at a controlled ambient temperature, pressure and humidity conditions. As a result, the models developed are valid for the specified conditions and a when the models are implemented, for instance, on a vehicle, the expectation is that the model works on a wide range of climatic conditions that the vehicle is exposed to, possibly conditions that were not experimented. Hence, an online adaptation to learn the behavior of the system at new/unfamiliar situations is required. Also, since the offline models are developed directly from experimental data, they may perform poorly in certain operating regions where the density of experimental data is low. As more data becomes available in such regions, an online mechanism can be used to adapt to such data. In addition, the engine produces high velocity streaming data; operating at about 2500 revolutions per minute, an in-cylinder pressure sensor can produce about 1.8 million data observations per day. It becomes infeasible to store this data for offline model development. Thus, an online learning framework that processes every data observation, updates the model and throws away the data is required for advanced engines like HCCI.

Online learning algorithms exist for linear and nonlinear models. For combustion engine applications, algorithms involving linear models are common in adaptive control. However, for a system like the HCCI engine, linear models may be insufficient to capture the complex dynamics and the authors showed that nonlinear identification models outperformed linear models, particularly for predicting several steps ahead in time \cite{vijay_asoc}. While numerous techniques for online learning do exist in machine learning literature, a complete survey is beyond the scope of this article. The recent paper on online sequential extreme learning machines (OS-ELM) \cite{oselm} surveys popular online learning algorithms in the context of classification and regression and develops an efficient algorithm based on recursive least squares. The OS-ELM algorithm seems to be the present state of the art for classification/regression problems achieving high generalization accuracies, global optimal solution and in quick time.

In spite of its known advantages, an over-parameterized ELM suffers from ill-conditioning problem when a recursive least squares type update is performed (as in OS-ELM). This sometimes results in poor regularization behavior \cite{illcond1,illcond2,illcond3,illcond4}, which leads to an unbounded growth of the model parameters and unbounded model predictions. If decisions are made as the model is updated (as in case of adaptive control for instance \cite{pred_adap}), it is vital for the parameter estimation to be stable so that model based decisions are valid. Hence a guarantee of stability and boundedness is of extreme importance. To address this issue, a stable online learning algorithm based on stochastic gradient descent is developed and stability is proved using Lyapunov stability theory. Although Lyapunov based approaches are popular in control theory, notable prior work for online learning include a Lyapunov approach applied for identification using radial basis function neural networks \cite{lyap_rbf} and GLO-MAP models \cite{glomap}. The parameter update in such methods involves complex gradient calculation in real time or first estimating a linear model and then estimating a nonlinear difference using orthonormal polynomial basis functions. The approach proposed in this paper aims to retain the simplicity and generalization power of ELM and OS-ELM algorithms, and introduce stability in parameter estimation so that such online models could be used for real-time control purposes.

The objective of this article is to develop a stable online learning algorithm for ELM models using stochastic gradients and apply to the HCCI engine modeling problem. The contributions of the paper are as follows. A novel online learning algorithm based on stochastic gradient descent for extreme learning machines is developed. The stability of parameter estimation for dynamic systems is proved using a Lyapunov stbility approach. The application of the stochastic gradient ELM algorithm to the complex HCCI engine identification is the first application (to our best knowledge) of online learning schemes to HCCI engines. This includes both the online state estimation problem as well as the online operating boundary estimation problem.

The remainder of the article is organized as follows. The ELM modeling approach is described in Section \ref{ELM_Sec} along with algorithm details on batch (offline) learning as well as the present state of the art - the OS-ELM. In Section \ref{sgELM_sec}, the stochastic gradient based ELM algorithm is derived along with stability proof. In Section \ref{hcci_sec}, the background on HCCI engine and experimentation are discussed. Sections \ref{cs1_sec} and \ref{cs2_sec} cover the discussions on the application of the SG-ELM algorithm on the two applications, followed by conclusions in Section \ref{concl}.

\section{Extreme Learning Machines} \label{ELM_Sec}
Extreme Learning Machine (ELM) is an emerging learning paradigm for multi-class classification and regression problems \cite{4Huang2005,huang12}. An advantage of the ELM method is that the training speed is extremely fast, thanks to the random assignment of input layer parameters which do not require adaptation to the data. In such a setup, the output layer parameters can be analytically determined using a least squares approach. Some of the attractive features of ELM \cite{4Huang2005} include the universal approximation capability of ELM, the convex optimization problem of ELM resulting in the smallest training error without getting trapped in local minima, closed form solution of ELM eliminating iterative training and better generalization capability of ELM \cite{huang12}.

Consider the following data set
\begin{equation}\label{tr_data}
\{(x_1,y_1),...,(x_N,y_N)\}\in \big(\mathcal{X},\mathcal{Y}\big),
\end{equation}
where $N$ denotes the number of training samples, $\mathcal{X}$ denotes the space of the input features and $\mathcal{Y}$ denotes labels whose nature differentiate the learning problem in hand. For instance, if $\mathcal{Y}$ takes integer values \{1,2,3,..\} then the problem is referred to as classification and if $\mathcal{Y}$ takes real values, it becomes a regression problem. ELMs are well suited for solving both regression and classification problems faster than state of the art algorithms \cite{huang12}. A further distinction could be made depending on the availability of training data during the learning process, as offline learning (or batch learning) and online learning (or sequential learning). Offline learning could make use of all training data simultaneously as all data is available to the algorithm. In addition, as the models are developed offline, efficient use of available computational resources could be made enabling offline algorithms to solve complex optimization problems. Typically, the accuracy of the modeling task takes priority over both computational demand and training time. On the other hand, situations where data is available as high velocity steams where it not feasible to store all data and make inference in quick time, or in situations where the inference is simultaneously made along with adaptation of model to incoming data, online learning is preferred. In an online learning setting, data is available one-by-one and needs to be processed with limited computational effort and storage. Further, inference is required to be made with each new available data along with the ones recorded in the past. In this work, the online setting is considered where a stable online learning algorithm is proposed that is compared with the offline approach and existing online learning method.

\subsection{Batch (Offline) ELM}
When the entire training data is available and a model is required to be learned using all the training data, batch learning is adopted. In this case, the ELM algorithm involves solving the following optimization problem
\begin{equation}\label{ELM_opti}
\min_{W}\left\{\|HW-Y\|^2+\lambda \|W\|^2\right\}
\end{equation}
\begin{equation}\label{hidden_output}
H^T=\psi(W_r^Tx(k)+b_r)\in\mathbb{R}^{n_h \times 1},
\end{equation}
where $\lambda$ represents the regularization coefficient, Y represents the vector of outputs or targets, $\psi$ represents the hidden layer activation function (sigmoidal, sinusoidal, radial basis etc \cite{huang12}) and $W_r\in\mathbb{R}^{n \times n_h}, W\in\mathbb{R}^{n_h \times y_d}$ represents the input and output layer parameters respectively. Here, $n$ represents the dimension of inputs $x(k)$, $n_h$ represents the number of hidden neurons of the ELM model, $H$ represents the hidden layer output matrix and $y_d$ represents the dimension of outputs $Y$. The matrix $W_r$ consists of randomly assigned elements that maps the input vector to a high dimensional feature space while $b_r\in\mathbb{R}^{n_h}$ is a bias component assigned in a random manner similar to $W_r$. The number of hidden neurons determines the expressive power of the transformed feature space. The elements can be assigned based on any continuous random distribution \cite{huang12} and remains fixed during the learning process. Hence the training reduces to a single step calculation given by equation \eqref{ELM_train}. The ELM decision hypothesis can be expressed as in equation \eqref{elm model} for classification and equation \eqref{elm model_reg} for regression. It should be noted that the hidden layer and the corresponding activation functions give a nonlinear mapping of the data, which if eliminated, becomes a linear least squares (Linear LS) model and is considered as one of the baseline models in this study.
\begin{equation}\label{ELM_train}
W^*=\left(H^TH + \lambda I \right)^{-1}H^TY
\end{equation}
\begin{equation}\label{elm model}
f(x)=\sgn\left(W^T[\psi(W_r^Tx+b_r)]\right).
\end{equation}
\begin{equation}\label{elm model_reg}
f(x)=W^T[\psi(W_r^Tx+b_r)]
\end{equation}

Since training involves a linear least squares solution with a convex objective function, the solution obtained by ELM is extremely fast and is a global optimum for the chosen $n_h$, $W_r$ and $b_r$. The above formulation for classification \eqref{elm model}, is not designed to handle imbalanced or skewed data sets. As a modification to weigh the minority class data more, a simple weighting method can be incorporated in the ELM objective function \eqref{ELM_opti} as
\begin{equation}\label{ELM_opti_weight}
\min_{W}\left\{(HW-Y)^T\Gamma(HW-Y)+\lambda W^T W\right\}
\end{equation}
\begin{center}
$\Gamma = \left [ \begin{array}{ccccc}
\gamma_1 & 0 & .& .& 0\\
0 & \gamma_2 & .& .& 0\\
. & . & .& .& 0\\
0 & 0 & .& .& \gamma_N\\
\end{array} \right ]$
\end{center}
\begin{equation}\label{weight_fac}
 \gamma_i =
  \begin{cases}
   1 &  \text{majority class data} \\
   r \times f_s  & \text{minority class data}
  \end{cases}
\end{equation}
where $\Gamma$ represents the weight matrix, $r$ represents the ratio of number of majority class data to number minority class data and $f_s$ represents a scaling factor to be tuned for a given data set \cite{vijay_open}. This results in the training step given by equation \eqref{ELM_train_weight} and the decision hypothesis takes the same form as in equation \eqref{elm model}:
\begin{equation}\label{ELM_train_weight}
W^*=\left(H^T \Gamma H + \lambda I \right)^{-1}H^T \Gamma Y.
\end{equation}

\subsection{Online Sequential ELM (OS-ELM)}
The OS-ELM \cite{oselm} is a recursive version of the batch ELM algorithm. This version of the algorithm is used for online learning purposes where data is processed one-by-one or chunk-by-chunk and the model parameters are updated after which the used data is not required to be stored. In this process, training involves two steps - initialization step and sequential learning step. During the initialization step, a set of data observations ($N_0$) are required to initialize the $H_0$ and $W_0$ by solving the following optimization problem
\begin{equation}\label{}
\min_{W_0}\left\{\|H_0W_0-Y_0\|^2+\lambda \|W_0\|^2\right\}
\end{equation}
\begin{equation}\label{}
H_0 = [g(W_r^T x_0+b_r)]^T \in \mathbb{R}^{N_0 \times n_h}.
\end{equation}
The solution $W_0$ is given by
\begin{equation}\label{}
W_0=K_0^{-1} H_0^T Y_0
\end{equation}
where $K_0 = H_0^T H_0+\lambda I$. Suppose given another new data $x_1$, the problem becomes
\begin{equation}\label{}
\min_{W_1}\left\|
\left[\begin{array}{cc}
H_0 \\
H_1 \end{array}\right]
W_1 -
\left[\begin{array}{cc}
Y_0\\
Y_1 \end{array}\right]
\right\|^2.
\end{equation}
The solution can be derived as
\begin{eqnarray*}
  W_1 &=& W_0 + K_1^{-1} H_1^T (Y_1 - H_1 W_0) \\
  K_1 &=& K_0 + H_1^T H_1.
\end{eqnarray*}
Based on the above, a generalized recursive algorithm for updating the least-squares solution can be computed as follows
\begin{equation}\label{P_eqn}
  M_{k+1} = M_k - M_k H^T_{k+1} (I + H_{k+1} M_k H_{K+1}^T )^{-1} H_{k+1} M_k
\end{equation}
\begin{equation}\label{W_eqn_ELM}
  W_{k+1} = W_k + M_{k+1} H^T_{k+1} (Y_{k+1} - H_{k+1} W_k)
\end{equation}
where $M$ represents the covariance of the parameter estimate.

\section{Stochastic Gradient Based ELM Algorithm}\label{sgELM_sec}
In this section, the proposed online learning algorithm using stochastic gradient descent (SGD) is developed for the extreme learning machine models for both classification and regression problems. SGD methods have been popular for several decades for performing online learning but with severe limitations on poor optimization and slow convergence rates. However, only recently, the asymptotic behavior of SGD methods has been analyzed indicating that SGD methods can be very powerful for learning large data sets \cite{sgd_bottou,sgd_exprate}. SGD based algorithms have been developed for Adaline networks, perceptron models, K-means, SVM and Lasso \cite{sgd_bottou}. In this work, the SGD algorithm is developed for extreme learning machines showing good potential for online learning of high velocity (streaming) data.

The justification of SGD based algorithms in machine learning can be briefly discussed as follows. In any learning problem, three types of errors are encountered, namely the approximation error, the estimation error and the optimization error \cite{sgd_bottou}, and the expected risk $E_{exp}(f)$ and the empirical risk $E_{emp}$ for a supervised learning problemd can be given by
\begin{eqnarray*}
E_{exp}(f) &=& \int l(f(x),y) dP (x,y) \\
E_{emp}(f) &=& \frac{1}{N} \sum_{i=1}^{N} l(f(x_i),y_i)
\end{eqnarray*}
Let $f^* = \text{argmin}_f E_{exp}(f)$ be the best possible prediction function. In practice, the prediction function is chosen from a family of parametric functions denoted by $\mathcal{F}$. Let $f^*_\mathcal{F} = \text{argmin}_{f \in \mathcal{F}} E_{exp}(f)$ be the best prediction function chosen from a parameterized family of functions $\mathcal{F}$. When a training data set becomes available, the empirical risk becomes a proxy for the expected risk for the learning problem \cite{vap95}. Let $\bar{f}^*_\mathcal{F} = \text{argmin}_{f \in \mathcal{F}} E_{emp}(f)$ be the solution that minimizes the empirical risk. However, the global solution is not typically obtained because of computational limitations and hence the solution of the learning problem is reduced to finding $\bar{f}_\mathcal{F} = \text{argmin}_{f \in \mathcal{F}} E_{emp}(f)$.

Using the above setup, the approximation error ($E_{app}$) is the error introduced in approximating the true function space with a family of functions $\mathcal{F}$, the estimation error ($E_{est}$) is the error introduced in optimizing over $E_{emp}(f)$ instead of $E_{exp}(f)$, the optimization error ($E_{opt}$) is the error induced as a result of stopping the optimization to $\bar{f}_\mathcal{F}$. The total error $E_{tot}$ can be expressed as
\begin{eqnarray*}
  E_{app} &=& E_{exp}(f^*)-E_{exp}(f^*_\mathcal{F}) \\
  E_{est} &=& E_{exp}(f^*_\mathcal{F})-E_{emp}(\bar{f}^*_\mathcal{F}) \\
  E_{opt} &=& E_{emp}(\bar{f}^*_\mathcal{F})-E_{emp}(\bar{f}_\mathcal{F}) \\
  E_{tot} &=& E_{app} + E_{est} + E_{opt}
\end{eqnarray*}

The following observations are taken from the asymptotic analysis of SGD algorithms \cite{sgd_bottou,Shalev-Shwartz_sgd}.
\begin{enumerate}
\item The empirical risk $E_{emp}(f)$ is only a surrogate for the expected risk $E_{exp}(f)$ and hence an increased effort to minimize $E_{opt}$ may not translate to better learning. In fact, if $E_{opt}$ is very low, there is a good chance that the prediction function will over-fit the training data.
\item SGD are worst optimization algorithms (in terms of reducing $E_{opt}$) but they minimize the expected risk relatively quickly. Therefore, in the large scale setup, when the limiting factor is computational time rather than the number of examples, SGD algorithms perform asymptotically better.
\item SGD results in a faster convergence when the loss function has strong convexity properties.
\end{enumerate}

The last observation is key in developing the algorithm based on ELM models. The ELM models have a squared loss function and when the hidden neurons are randomly assigned and fixed, the training translates to solving a convex optimization problem. Hence the ELM model can be a good candidate to perform SGD type learning and hence the motivation for this study. The SGD based algorithm can be derived for the ELM models as follows.

\subsection{Algorithm Formulation} \label{algo_derive}
Let ($x_i,y_i$) where $i=1,2,..N$ be the streaming data in consideration. The data can be considered to be available to the algorithm from a one-by-one continuous stream or artificially sampled one-by-one from a very large data set. Let the ELM empirical risk be defined as follows
\begin{eqnarray}\label{}
\nonumber J(W) &=& \min_W \frac{1}{2} \sum_{i=1}^{N} \| y_i - \phi_i^T W \|^2 \\
\nonumber     &=& \min_W \left\{ \frac{1}{2} \| y_1 - \phi_1^T W \|^2 + . . + \frac{1}{2} \| y_N - \phi_N^T W \|^2 \right\} \\
     &=& \min_W \left\{ J_1(W) + J_2(W) + . . + J_N(W) \right\}.
\end{eqnarray}
where $W \in \mathbb{R}^{n_h \times y_d}$, $y_i \in \mathbb{R}^{1 \times y_d}$ $\phi \in \mathbb{R}^{n_h \times y_d}$ is the hidden layer output (see $H^T$ in equation \eqref{hidden_output}). If an error $e_i \in \mathbb{R}^{1 \times y_d}$ can be defined as $(y_i - \phi_i^T W)$, the learning objective for a data observation $i$ can be given by
\begin{eqnarray}
\nonumber  J_i(W) &=& \frac{1}{2} e_i^T e_i \\
\nonumber  &=& \frac{1}{2} (y_i-\phi_i^T W)^T (y_i-\phi_i^T W) \\
\nonumber  &=& \frac{1}{2}y_i^T y_i + \frac{1}{2} W^T \phi_i \phi_i^T W -y_i^T \phi_i^T W \\
\nonumber  \frac{\partial J_i}{\partial W} &=& \phi_i \phi_i^T W - \phi_i y_i = \phi_i (\phi_i^T W - y_i) \\
           &=& -\phi_i e_i.
\end{eqnarray}
In a regular gradient descent (GD) algorithm, the gradient of $J(W)$ is used to update the model parameters as follows.
\begin{eqnarray}\label{GD_update}
\nonumber  \frac{\partial J}{\partial W} &=& \frac{\partial J_1}{\partial W} + \frac{\partial J_2}{\partial W} + .. + \frac{\partial J_N}{\partial W} \\
\nonumber  \Rightarrow \frac{\partial J}{\partial W} &=& - \phi_1 e_1 - \phi_2 e_2 - .. - \phi_N e_N \\
\nonumber  W_{k+1} &=& W_k-\Gamma_{SG} \frac{\partial J}{\partial W} \\
         &=& W_k + \Gamma_{SG} (\phi_1 e_1) + .. + \Gamma_{SG} (\phi_N e_N)
\end{eqnarray}
where $k$ is the iteration count, $\Gamma_{SG} \in \\mathbb{R}^{n_h \times n_h}$ represents the step size or update gain matrix for the GD algorithm.

It can be seen from equation \eqref{GD_update} that the parameter matrix $W$ is updated based on gradients calculated from all the available examples. If the number of data observations is large, the gradient calculation can take enormous computational effort. The stochastic gradient descent algorithm considers one example at a time and updates $W$ based on gradients calculated from ($x_i,y_i$) as shown in
\begin{equation} \label{SGD update}
W_{i+1} = W_i + \Gamma_{SG} (\phi_i e_i).
\end{equation}
From equation \eqref{GD_update}, it is clear that the optimal $W$ is a function of gradients calculated from all the examples. As a result, as more data becomes available, $W$ converges close to its optimal value in SGD algorithm. Processing data one-by-one significantly reduces the computational requirement and the algorithm is scalable to large data sets. More importantly, for the online learning of HCCI engine dynamic considered in this work, the SGD algorithm becomes a strong candidate.

In order to handle class imbalance learning, the algorithm in \eqref{SGD update} can be modified by weighting the minority class data more. The modified algorithm can be expressed as
\begin{equation} \label{SGD update_cil}
W_{i+1} = W_i + \Gamma_{imb} \Gamma_{SG} (\phi_i e_i)
\end{equation}
where $\Gamma_{imb}=r \times f_s$, $r$ and $f_s$ represent the imbalance ratio (a running count of majority class data to minority class data until that instant) and the scaling factor that needs to be tuned to obtain tradeoffs between high false positives and missed detections for a given application.

\subsection{Stability Analysis} \label{stab_sec}
The stability analysis of the SGD based ELM algorithm can be derived as follows. The ELM structure makes the analysis simple and similar to that of a linear gradient based algorithm \cite{jingsun}.

The instantaneous prediction error $e_i$ (Here the error $e$ and output $y$ are transposed as opposed to their previous definition in Section \ref{algo_derive} for ease of derivations) can be expressed in terms of parametric error ($\tilde{W}=W_*-W$) as
\begin{eqnarray}\label{sg_err_par_err}
\nonumber e_i &=& y_i-W^T \phi_i \\
\nonumber &=& W_*^T \phi_i-W^T \phi_i \\
&=& \tilde{W}^T\phi_i
\end{eqnarray}
where $W_*$ represents true model parameters. Further, the parametric error dynamics can be obtained as follows.
\begin{eqnarray}
\nonumber \tilde{W}_{i+1} &=& W_* - W_{i+1} \\
\nonumber &=& W_*-W_i-\Gamma_{SG}\phi_i e_i^T \\
&=& \tilde{W}_i-\Gamma_{SG}\phi_i e_i^T
\end{eqnarray}

Consider the following positive definite, decrescent and radially unbounded \cite{jingsun} Lyapunov function $V$
\begin{equation}\label{}
V(\tilde{W})=tr(\tilde{W}^T\Gamma_{SG}^{-1}\tilde{W})
\end{equation}
where $tr$ represents the trace of a matrix.
\begin{eqnarray}\label{sg_lyap}
\nonumber   \Delta V(\tilde{W}_i)&=& V(\tilde{W}_{i+1}) - V(\tilde{W}_i) \\
\nonumber   &=& tr(\tilde{W}_{i+1}^T\Gamma_{SG}^{-1}\tilde{W}_{i+1}) - tr(\tilde{W}_i^T\Gamma_{SG}^{-1}\tilde{W}_i) \\
\nonumber   &=& tr((\tilde{W}_i-\Gamma_{SG}\phi_i e_i^T)^T\Gamma_{SG}^{-1}(\tilde{W}_i-\Gamma_{SG}\phi_i e_i^T)) \\
\nonumber   && - tr(\tilde{W}_i^T\Gamma_{SG}^{-1}\tilde{W}_i) \\
\nonumber   &=& tr( -2 \tilde{W}_i^T \phi_i e_i^T + e_i \phi_i^T \Gamma_{SG} \phi_i e_i^T) \\
\nonumber    &=& tr( -2 e_i e_i^T + e_i \phi_i^T \Gamma_{SG} \phi_i e_i^T) \\
\nonumber    &=&  -2 e_i^T e_i + e_i^T e_i \phi_i^T \Gamma_{SG} \phi_i \\
\nonumber    &=&  -2 e_i^T e_i + e_i^T \phi_i^T \Gamma_{SG} \phi_i e_i \\
   &=& - e_i^T M_{SG} e_i
\end{eqnarray}
where $M_{SG} = 2-\phi_i^T \Gamma_{SG} \phi_i$. It can be seen that $V_{i+1} - V_i \leq 0$ if $M_{SG}>0$ or $2-\phi_i^T \Gamma_{SG} \phi_i > 0$ or
\begin{equation}\label{gamma_sg_cond1}
0<\lambda_{max}(\Gamma_{SG})<2
\end{equation}
When \eqref{gamma_sg_cond1} is satisfied, $V(\tilde{W}) \geq 0$ is non-increasing in $i$ and the limit
\begin{equation}\label{}
\lim_{k \rightarrow \infty}V(\tilde{W})=V_\infty
\end{equation}
exists. From \eqref{sg_lyap},
\begin{eqnarray}
\nonumber V_{i+1}-V_i &=& - e_i^T M_{SG} e_i \\
\nonumber \sum_{i=0}^\infty (V_{i+1}-V_i) &=& - \sum_{i=0}^\infty e_i^T M_{SG} e_i \\
\Rightarrow \sum_{i=0}^\infty e_i^T M_{SG} e_i &=& V(0)-V_\infty < \infty \\
\end{eqnarray}
Also,
\begin{equation}\label{}
\sum_{i=0}^\infty e_i^T I e_i \leq \sum_{i=0}^\infty e_i^T M_{SG} e_i < \infty
\end{equation}
when $M_{SG}>I$ or when
\begin{equation}\label{gamma_sg_cond2}
\lambda_{max}(\Gamma_{SG})<1.
\end{equation}
Hence, when \eqref{gamma_sg_cond2} is satisfied, $e_i \in \mathbf{L}_2$. From \eqref{SGD update}, $(W_{i+1} - W_i)\in\mathbf{L}_2 \cap \mathbf{L}_\infty$. Using discrete time Barbalat's lemma \cite{disc_barb_lemma},
\begin{eqnarray}
\lim_{i \rightarrow \infty} e_i &=& 0 \\
\lim_{i \rightarrow \infty} W_{i+1} &=& W_i
\end{eqnarray}

Hence, the SGD learning law in \eqref{SGD update} guarantees that the estimated output $\hat{y}_i$ converges to the actual output $y_i$ and the model parameters $W$ converge to some constant values. The parameters converge to the true parameters $W_*$ only under conditions of persistence of excitation \cite{jingsun} in input signals of the system (amplitude and frequency richness of $x$). Further, using boundedness of $V_i$, $e_i \in\mathbf{L}_\infty$  which guarantees that the online model predictions are bounded as long as the system output is bounded. As the error between the true model and the estimation model converges to zero, the estimation model becomes a one-step ahead predictive model of the nonlinear system. The evaluation of the SG-ELM algorithm is performed using application to a complex HCCI engine identification problem.

\section{Homogeneous Charge Compression Ignition Engine} \label{hcci_sec}
\begin{figure*}[hptb]
      \centering
      \includegraphics[scale=0.7]{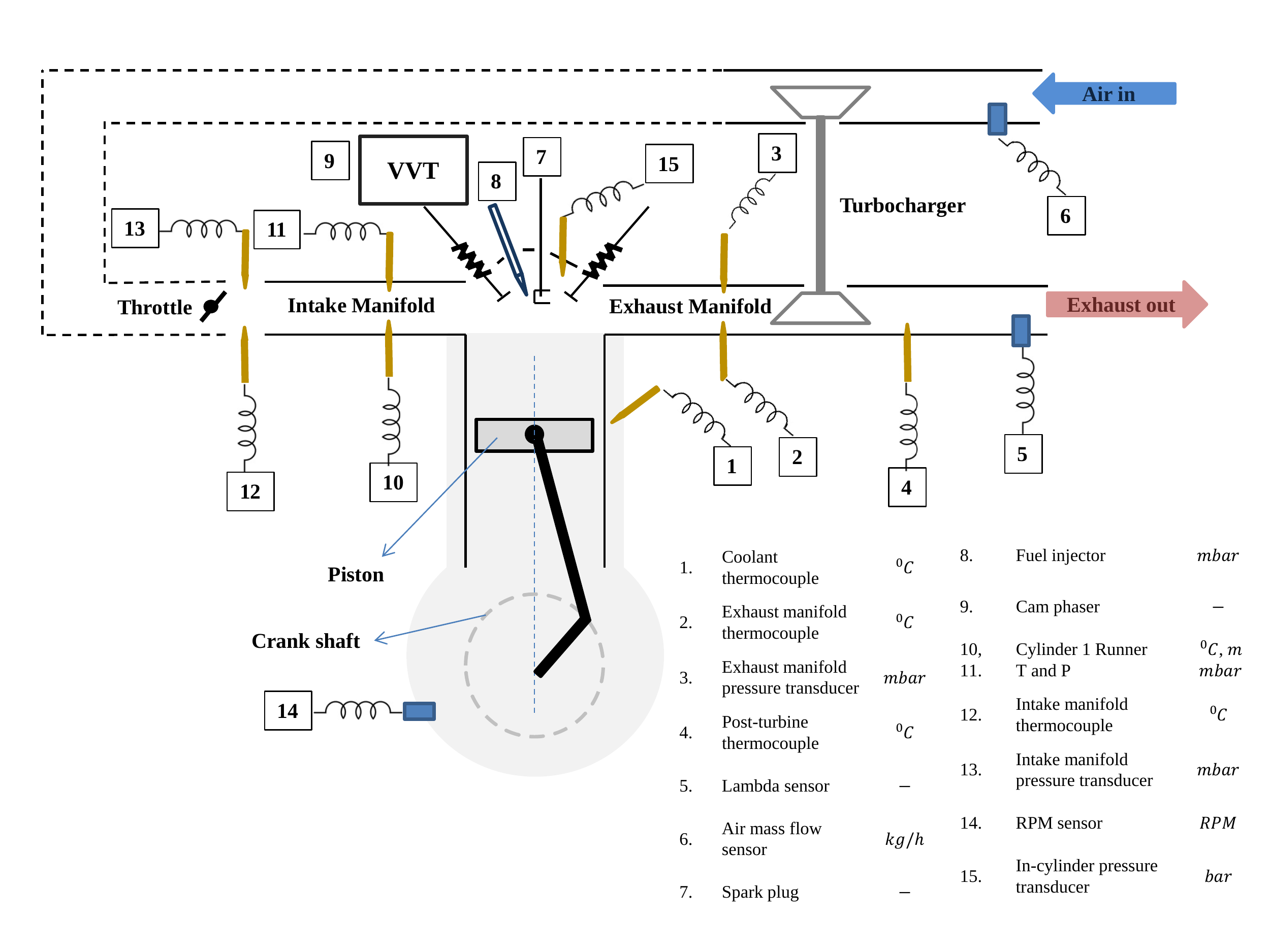}
      \caption{A schematic of the HCCI engine setup and instrumentation (only relevant instrumentation shown).}
      \label{schematic}
\end{figure*}

The algorithms discussed in Section \ref{ELM_Sec} are applied to streaming sensory data from a gasoline HCCI engine for demonstrating an online learning framework for HCCI engine modeling. The engine specifications are listed in Table \ref{specstable} \cite{vijay_asoc}. A schematic of the experimental setup and instrumentation is shown in Fig. \ref{schematic}. HCCI is achieved by auto-ignition of the gas mixture in the cylinder. The fuel is injected early in the intake stroke and given sufficient time to mix with air forming a homogeneous mixture. A large fraction of exhaust gas from the previous cycle is retained to elevate the temperature and hence the reaction rates of the fuel and air mixture. The variable valve timing capability of the engine enables trapping suitable quantities of exhaust gas in the cylinder.

\begin{table}[hptb]
\caption{Specifications of the experimental HCCI engine}
\label{specstable}
\footnotesize
\begin{center}
\begin{tabular}[c]{|c|c|}
\hline
Engine Type & 4-stroke In-line\\
\hline
Fuel & Gasoline\\
\hline
Displacement & 2.0 L\\
\hline
Bore/Stroke & 86/86 mm\\
\hline
Compression Ratio & 11.25:1\\
\hline
Injection Type & Direct Injection\\
\hline
 & Variable Valve Timing with \\
&  hydraulic cam phaser having\\
Valvetrain  & 119 degree constant duration \\
 & defined at 0.25mm lift, 3.5mm peak \\
 & lift and 50 degree crank angle \\
 & phasing authority \\
 \hline
HCCI strategy & Exhaust recompression \\
& using negative valve overlap\\
\hline
\end{tabular}
\end{center}
\end{table}

The engine can be controlled using precalculated inputs such as injected fuel mass (FM in mg/cyc), crank angle at intake valve opening (IVO), crank angle at exhaust valve closing (EVC), crank angle at start of fuel injection (SOI). The valve events are measured in degrees after exhaust top dead center (deg eTDC) while SOI is measured in degrees after combustion top dead center (deg cTDC). Other important physical variables that influence the performance of HCCI combustion include intake manifold temperature $T_{in}$, intake manifold pressure $P_{in}$, mass flow rate of air at intake $\dot{m}_{in}$, exhaust gas temperature $T_{ex}$, exhaust manifold pressure $P_{ex}$, coolant temperature $T_{c}$, fuel to air ratio (FA) etc. The engine performance metrics are given by combustion phasing indicated by the crank angle at 50\% mass fraction burned (CA50), combustion work output given by net indicated mean effective pressure (NMEP, sometimes abbreviated as IMEP). The combustion features calculated using in-cylinder pressure such as CA50, NMEP are determined from the high speed in-cylinder pressure measurements. For further reading on HCCI combustion and related variables, please refer \cite{hcci_book}.

\subsection{Experiment Design}
In order to identify both models for HCCI state variables as well as models for dynamic operating boundary in transient operation, appropriate experiment design to obtain transient data from the engine is required. The modeled variables such as engine states and operating envelope are dynamic variables and in order to capture both transient and steady state behavior, a set of dynamic experiments is conducted at constant rotational speeds and naturally aspirated conditions (no supercharging/turbocharging) by varying FM, IVO, EVC and SOI in a uniformly random manner. Every input step involves the engine making a transition between two set conditions and the transition (transients or dynamics) is recorded as temporal data. In order to capture several such transients, an amplitude modulated pseudo-random binary sequence (A-PRBS) has been used to design the excitation signals. A-PRBS enables exciting the engine at different amplitudes and frequencies suitable for the identification problem considered in this work. The data is sampled using the AVL Indiset acquisition system where in-cylinder pressure is sensed every crank angle using which the combustion features NMEP, CA50 are determined on a per-combustion cycle basis. More details on HCCI combustion and experiments can be found in \cite{vijay_asoc,vijay_springer,vijay_open}

\subsection{HCCI Instabilities}\label{instab_sec}
\begin{figure}[]
      \centering
      \includegraphics[scale=0.37]{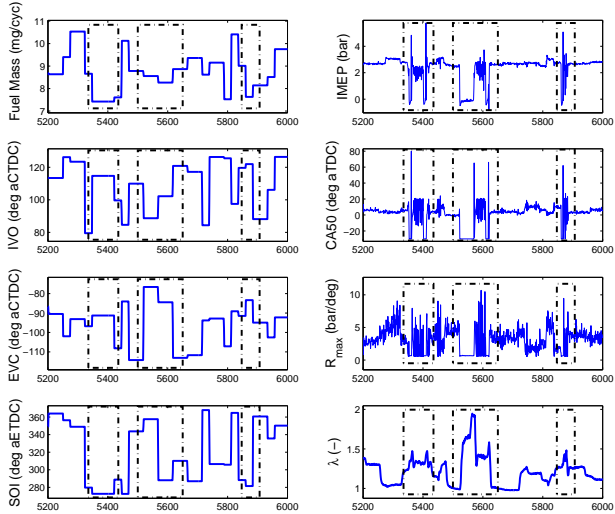}
      \caption{A subset of the HCCI engine experimental data showing A-PRBS inputs and engine outputs. The misfire regions are shown in dotted rectangles. The data is indexed by combustion cycles.}
      \label{misfire fig}
\end{figure}

A subset of the data collected from the engine is shown in Fig. \ref{misfire fig} where it can be observed that for some combinations of the inputs (left figures), the HCCI engine misfires (seen in the right figures where NMEP drops below 0 bar). HCCI operation is limited by several phenomena that lead to undesirable engine behavior. As described in \cite{george}, the HCCI operating range is conceptually constrained to a small region of permissible unburned (pre-combustion) and burned (post-combustion) charge temperature states. As previously noted, sufficiently high unburned gas temperatures are required to achieve ignition in the HCCI operating range without which complete misfire will occur. If the resulting combustion cannot achieve sufficiently high burned gas temperatures, commonly occurring in conditions with low fuel to diluent ratios or late combustion phasing, various degrees of quenching can occur resulting in reduced work output and increased hydrocarbon and carbon monoxide emissions. Under some conditions, this may lead to high cyclic variation due to the positive feedback loop existing through the trapped residual gas \cite{misf1,misf2}. Operation with high burned gas temperature, although stable and commonly reached at higher fueling rates where the fuel to diluent ratio is also high, yields high heat release and thus pressure rise rates that may pose challenges for engine noise and durability constraints. A discussion of the temperatures at which these phenomena occur may be found in \cite{george}.

\subsection{Learning The HCCI Engine Data} \label{data_formulation_sec}
In the HCCI modeling problem, both the inputs and the outputs of the engine are available as sensor measurements and hence supervised learning can be employed. The HCCI engine is a nonlinear dynamic system and sensor measurements represent discrete time sequences. The input-output behavior can be modeled using a nonlinear auto regressive model with exogenous input (NARX) \cite{Nelles17} as follows
\begin{multline}\label{NARX eqn}
y(k)=f_{NARX}[u(k-1),..,u(k-n_u), \\
            y(k-1),..,y(k-n_y)]
\end{multline}
where $u(k)\in\mathbb{R}^{u_d}$ and $y(k)\in\mathbb{R}^{y_d}$ represent the inputs and outputs of the system respectively, $k$ represents the discrete time index, $f_{NARX}(.)$ represents the nonlinear function mapping specified by the model, $n_u$, $n_y$ represent the number of past input and output samples required (order of the system) while $u_d$ and $y_d$ represent the dimension of inputs and outputs respectively. Let $x$ represent the augmented input vector obtained by appending the input and output measurements from the system.
\begin{equation}\label{inp feature}
x=[u(k-1),..,u(k-n_u),y(k-1),..,y(k-n_y)]^T
\end{equation}
The input measurement sequence can be converted to the form of training data
\begin{equation}\label{}
\{(x_1,y_1),...,(x_N,y_N)\}\in \big(\mathcal{X},\mathcal{Y}\big)
\end{equation}
where $N$ denotes the number of training samples, $\mathcal{X}$ denotes the space of the input features (Here $\mathcal{X} = \mathbb{R}^{u_dn_u+y_dn_y}$ and $\mathcal{Y} = \mathbb{R}$ for regression and $\mathcal{Y} =\{+1,-1\}$ for a binary classification). The above conversion of system measurements to training data is a natural definition for a series-parallel model architecture and the models can be used for a one-step ahead prediction (OSAP) i.e., given a set of measurements until time index $k$, the model predicts the output at time $k+1$ (see equation \eqref{NARX pred_sp}). A parallel architecture on the other hand can be used to perform multiple step ahead predictions (MSAP) by feeding back the predictions of the OSAP model in a recurrent manner (see equation \eqref{NARX pred_p}). The series-parallel and parallel architectures are well explained in \cite{narendra}.
\begin{multline}\label{NARX pred_sp}
\hat{y}(k+1) = \hat{f}_{NARX}[u(k),..,u(k-n_u+1),y(k),\\
                ..,y(k-n_y+1)]
\end{multline}
\begin{multline}\label{NARX pred_p}
\hat{y}(k+n_{pred}) = \hat{f}_{NARX}[u(k+n_{pred}-1),..,u(k-n_u+n_{pred}), \\
                        \hat{y}(k+n_{pred}-1),..,\hat{y}(k-n_y+n_{pred})]
\end{multline}
The OSAP model is used for training as existing simple training algorithms can be used and once the model becomes accurate for OSAP, it can be converted to a MSAP model in a straightforward manner. The MSAP model can be used for making long term predictions useful for predictive control \cite{Johansson2010,auto_mpc,vijay_control}.

\section{Application Case Study 1: Online regression learning for system identification of an HCCI Engine.} \label{cs1_sec}
As mentioned earlier, a key requirement for model based control of the HCCI engine is the ability to accurately predict the engine state variables for several operating cycles ahead of time, so that a control action with a known impact can be applied to the engine. The state variables of an engine are the fundamental quantities that represent the engine's state of operation. As a consequence, these variables also influence the performance of the engine such as fuel efficiency, emissions and stability, and are required to be monitored/regulated. In this section, the NMEP and CA50 are considered indicative of engine state variables and are estimated based on control inputs alone, so that the resulting models can be used for predictive control. This section details the experiments, model training and validation of the identified models.

For the HCCI control oriented modeling, an online regression learning framework is developed. In contrast to the existing linear system identification \cite{Johansson2010}, a nonlinear identification is employed. Typical features of nonlinear identification such as slow convergence and complex parameter update make existing methods practically unsuitable for complex systems. In this work, these shortcomings are eliminated making the approach suitable for the complex HCCI engine problem in hand.

\subsection{Model Structure and Evaluation Metric}
For the purpose of demonstration, the variables NMEP and CA50 are considered as outputs whereas the control variables such as fueling (FM), exhaust valve closing (EVC) and fuel injection timing (SOI) are considered inputs. Transient data from the HCCI engine at a constant speed of 1800 RPM and naturally aspirated conditions is used. A NARX model as shown in section \ref{data_formulation_sec} is considered where $u=[FM \quad EVC \quad SOI]^T$ and $y=[NMEP \quad CA50]^T$, $n_u$ and $n_y$ chosen as 1 (tuned by trial and error). The nonlinear model approximating $f_{NARX}$ is initialized to an extreme learning machine model with random input layer weights and random values for the covariance matrices and output layer weights. Four different models are considered including the state of the art OS-ELM algorithm, the proposed SG-ELM algorithm, a baseline offline (batch) ELM (O-ELM) and a baseline linear system identification model. The purpose of the baseline offline ELM algorithm is to evaluate the efficiency of the online learning models in learning the HCCI behavior completely as an offline ELM model would do. The offline ELM model is expected to produce an accurate model as it has sufficient time, computation and utilization of all training data simultaneously to learn the HCCI behavior sufficiently well. The purpose of the linear baseline model is to justify the use of a nonlinear model for HCCI dynamics.

All the nonlinear models consist of 100 hidden units with fixed randomized input layer parameters. About 11000 cycles of data is considered one-by-one as it is sampled by the engine ECU and model parameters updated in a sequential manner. After the training phase, the parameter update is switched off and the models are evaluated for the next 5100 cycles of data for one step ahead predictions. Further, to evaluate if the learned models represent the actual HCCI dynamics, the multi-step ahead prediction of the models are compared using about 600 cycles of data. It should be noted that both the one-step ahead and multi-step ahead evaluations were done using data unseen during the training phase.

The parameters of each of the models are tuned to accurately represent the given dataset. As recommended by OS-ELM \cite{oselm}, about 800 cycles of data was used for initializing the output layer parameters $W_0$ and covariance matrix $M_0$ (see equations \eqref{P_eqn} and \eqref{W_eqn_ELM}). The initialization was performed using the batch ELM algorithm \cite{huang12}. In order to have a fair comparison, the $W_0$ is used as an initial condition for both OS-ELM and SG-ELM. The only parameter of SG-ELM, namely the gradient step size was tuned to be $\Gamma_{SG} = 0.0008 \quad I_{100}$ for best accuracy. This was determined using trial and error and the value of $\Gamma_{SG}$ had a significant impact on the prediction accuracy. A detailed analysis on the robustness of $\Gamma_{SG}$ is outside the scope of this paper and will be considered for future.

The performance of the models are measured using normalized root mean squared error (RMSE) given by
\begin{equation}\label{}
RMSE=\sqrt{\frac{1}{n}\sum_{i=1}^{n}\sum_{j=1}^{y_d}(y_j^i-\hat{y}_j^i)^2}
\end{equation}
where both $y_j^i$ and $\hat{y}_j^i$ are normalized to lie between -1 and +1.

\subsection{Results and Discussion}
\begin{figure*}[]
      \centering
      \includegraphics[scale=1]{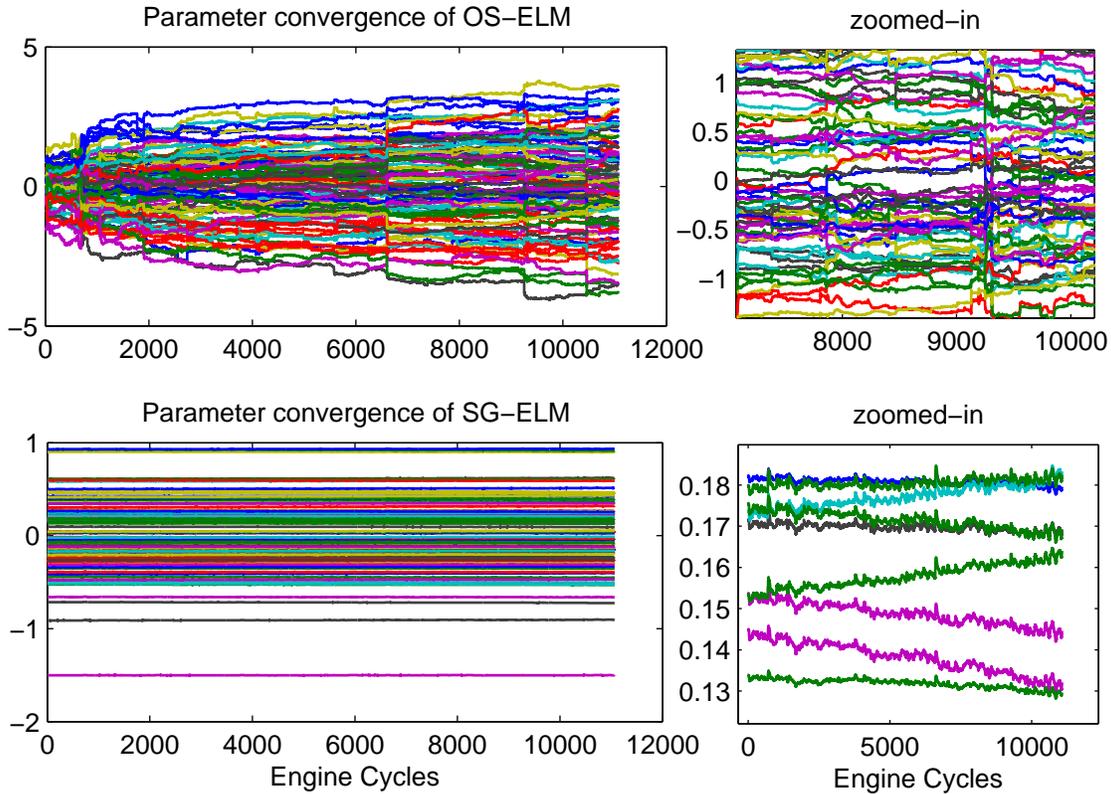}
      \caption{Comparison of parameter evolution for the OS-ELM and SG-ELM algorithms during online learning. A zoomed-in plot shows that the parameter update for OS-ELM is more aggressive compared to SG-ELM. Both OS-ELM and SG-ELM are initialized to the same parameters. The less aggressive and slow variation of the SG-ELM parameters along with stability bounds result in better regularization compared to OS-ELM.}
      \label{par_evol}
\end{figure*}

On performing online learning, it can be observed from Fig. \ref{par_evol} that the parameters of OS-ELM grow more aggressively as compared to the SG-ELM. In spite of both models having the same initial conditions, the step size parameter $\Gamma_{SG}$ for SG-ELM gives additional control over the parameter growth and keep them bounded as proved in section \ref{stab_sec}. On the other hand, OS-ELM doesn't have any control over the parameter evolution. It is governed by the evolution of the co-variance matrix $M$ \eqref{P_eqn}. It is expected that the co-variance matrix $M$ would add stability to the parameter evolution but in practice, it tends to be more aggressive leading to potential instabilities as reported by \cite{illcond1,illcond2,illcond3,illcond4}. As a consequence, the parameter values for SG-ELM remain small compared to the OS-ELM (the norm of estimated parameters for OS-ELM is 16.64 and SG-ELM is 3.71). This has a significant implication in the statistical learning theory \cite{vapnik}. A small norm of model parameters implies a simpler model which results in good generalization. Although this effect is slightly reflected in the results summarized in prediction results summarized in Table \ref{reg_res_table} (see MSAP RMSE for SG-ELM being the lowest), it is not significantly better for this problem possibly because of incomplete convergence. The value of $\Gamma_{SG}$ has to be tuned correctly along with sufficient training data in order to ensure parameter convergence. Ultimately, the online learning mechanism is aimed to run along with the engine and hence the slow convergence may not be an issue in a vehicle application.

\begin{table}[htbp]
  \centering
  \caption{Performance comparison of OS-ELM and SG-ELM for the HCCI online regression learning problem. A baseline linear model and an offline trained ELM model (O-ELM) are also included for comparison.}
    \begin{tabular}{cccc}
    \hline
          & Training & OSAP  & MSAP \\
          & Time in s & RMSE  & RMSE \\
    \hline
    Linear  & 0.3523 & 0.2004 & 0.1664 \\
    OS-ELM & 3.3812 & \textbf{0.0957} & 0.1024 \\
    SG-ELM & \textbf{0.7269} & 0.1047 & \textbf{0.0939} \\
    O-ELM & -     & 0.1015 & 0.1003 \\
    \hline
    \end{tabular}%
  \label{reg_res_table}%
\end{table}%

The prediction results as well as training time for the online models are compared in Table \ref{reg_res_table}. It can be observed that the computational time for SG-ELM is significantly less (about 4.6 times) compared to OS-ELM indicating the SG-ELM features a faster learning. The reduction in computation is expected to be more pronounced as the dimension and complexity of the data increase. It could be seen from Table \ref{reg_res_table} that the one-step ahead prediction accuracies (OSAP RMSE) of the nonlinear models are similar with OS-ELM winning marginally. On the other hand, the multi-step prediction accuracies (MSAP RMSE) are similar for the nonlinear models with SG-ELM performing marginally better. The MSAP accuracy reflect the generalization performance of the model and is more crucial for the modeling problem as the models ultimately feed its prediction to a predictive control framework that requires accurate and robust predictions of the engine several steps ahead of time. From our understanding on model complexity and generalization error, a model that is less complex (indicated by minimum norm of parameters \cite{huang12,vap95}) tend to generalize better, which is again demonstrated by SG-ELM. The performance of the linear baseline model is significantly low compared to the nonlinear models justifying adopting a nonlinear identification for the HCCI engine problem.

\begin{figure*}[]
      \centering
      \begin{tabular}{cc}
      \subfloat[OS-ELM MSAP Prediction]{\label{1800_pred_1}\includegraphics[scale=0.6]{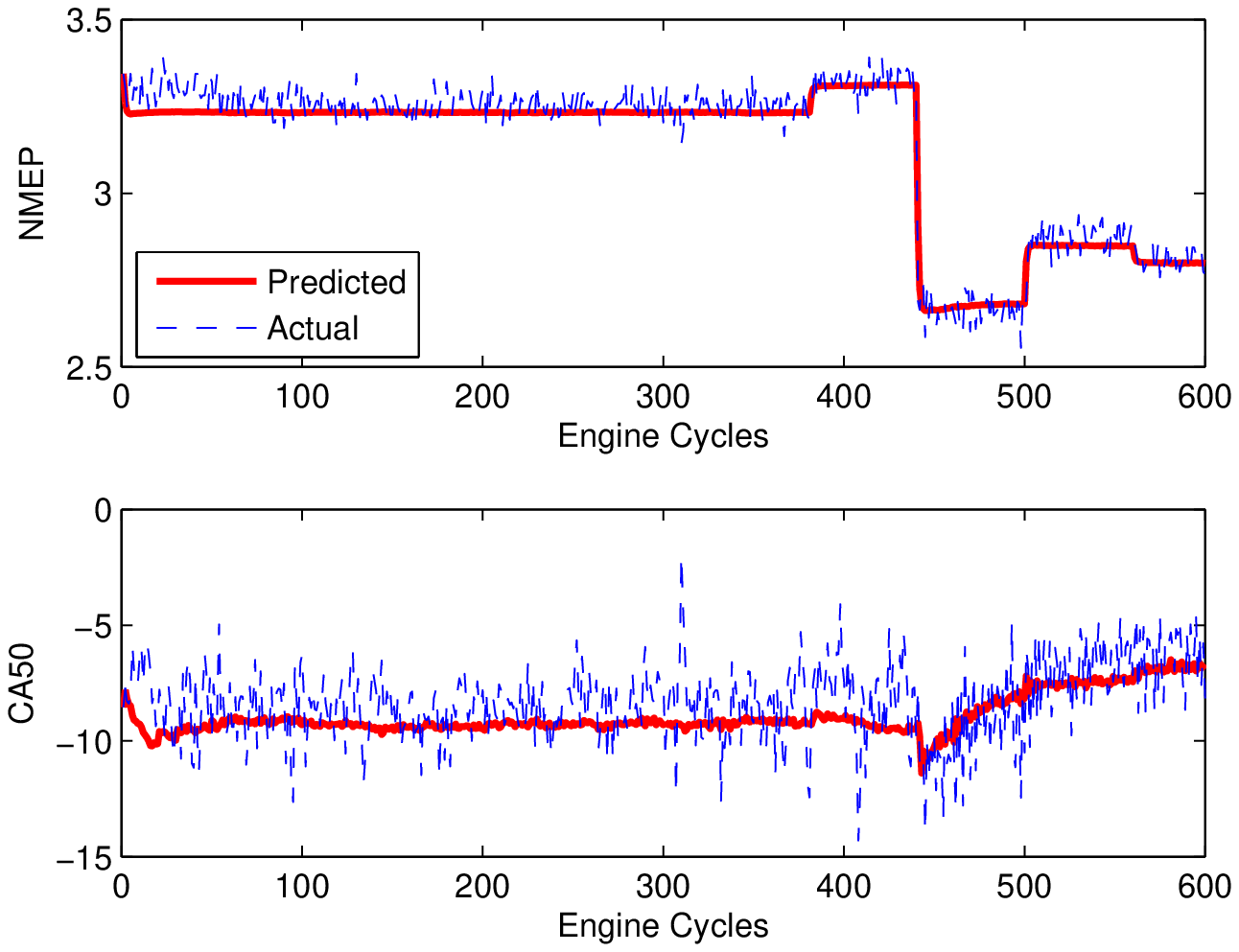}}
      &      \subfloat[SG-ELM MSAP Prediction]{\label{1800_pred_2}\includegraphics[scale=0.6]{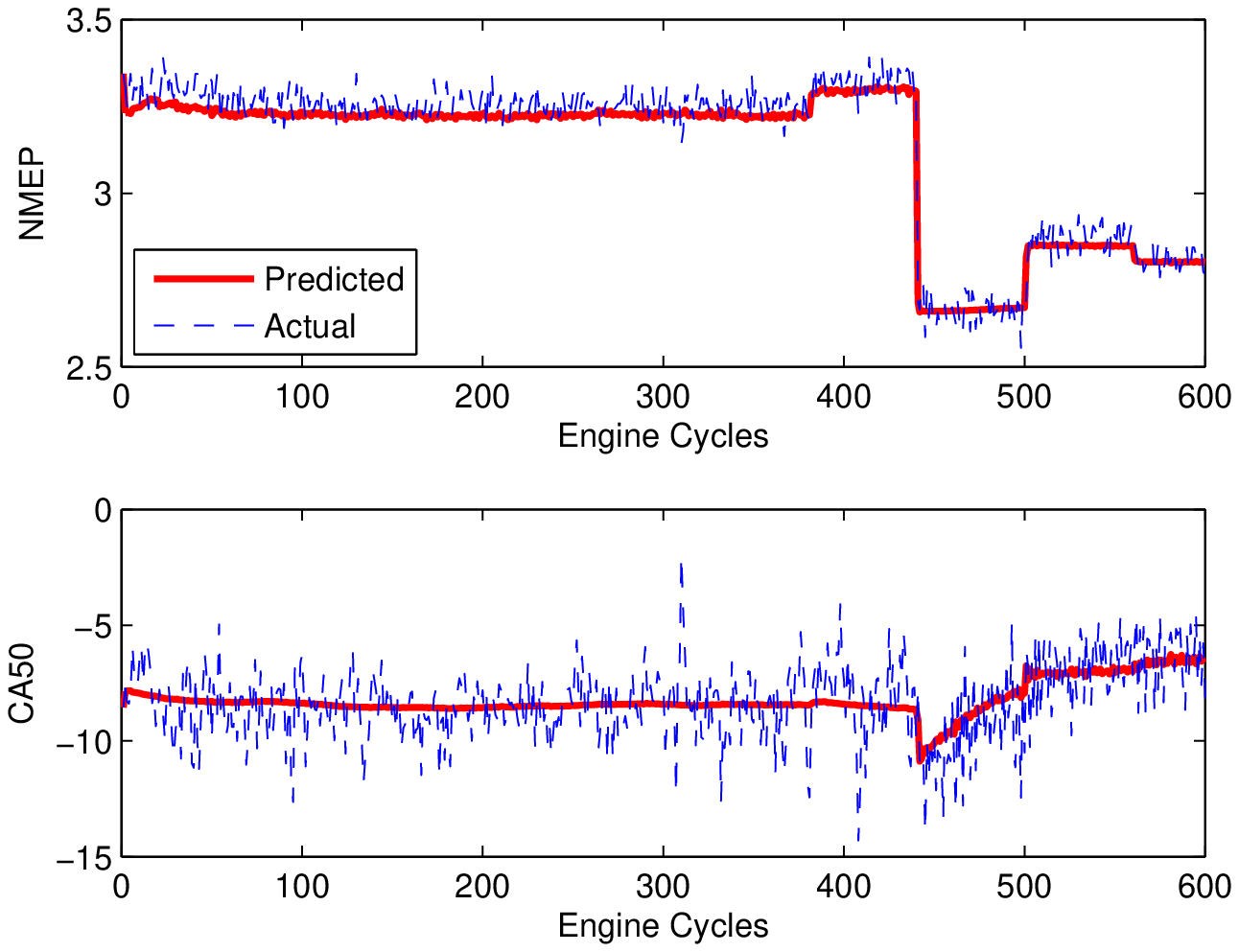}}\\
      \subfloat[O-ELM MSAP Prediction]{\label{1800_pred_3}\includegraphics[scale=0.6]{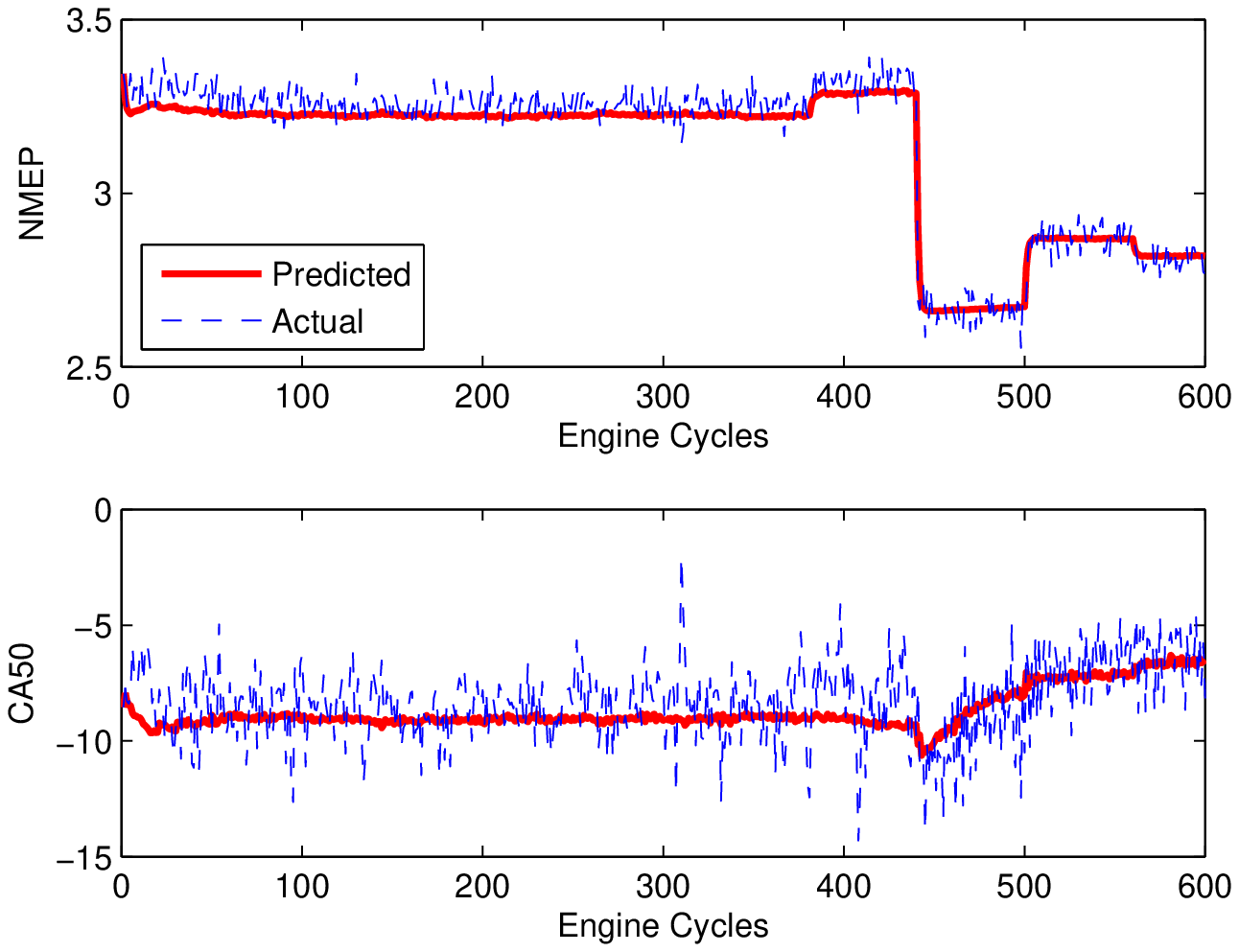}}
      &      \subfloat[Linear MSAP Prediction]{\label{1800_pred_4}\includegraphics[scale=0.6]{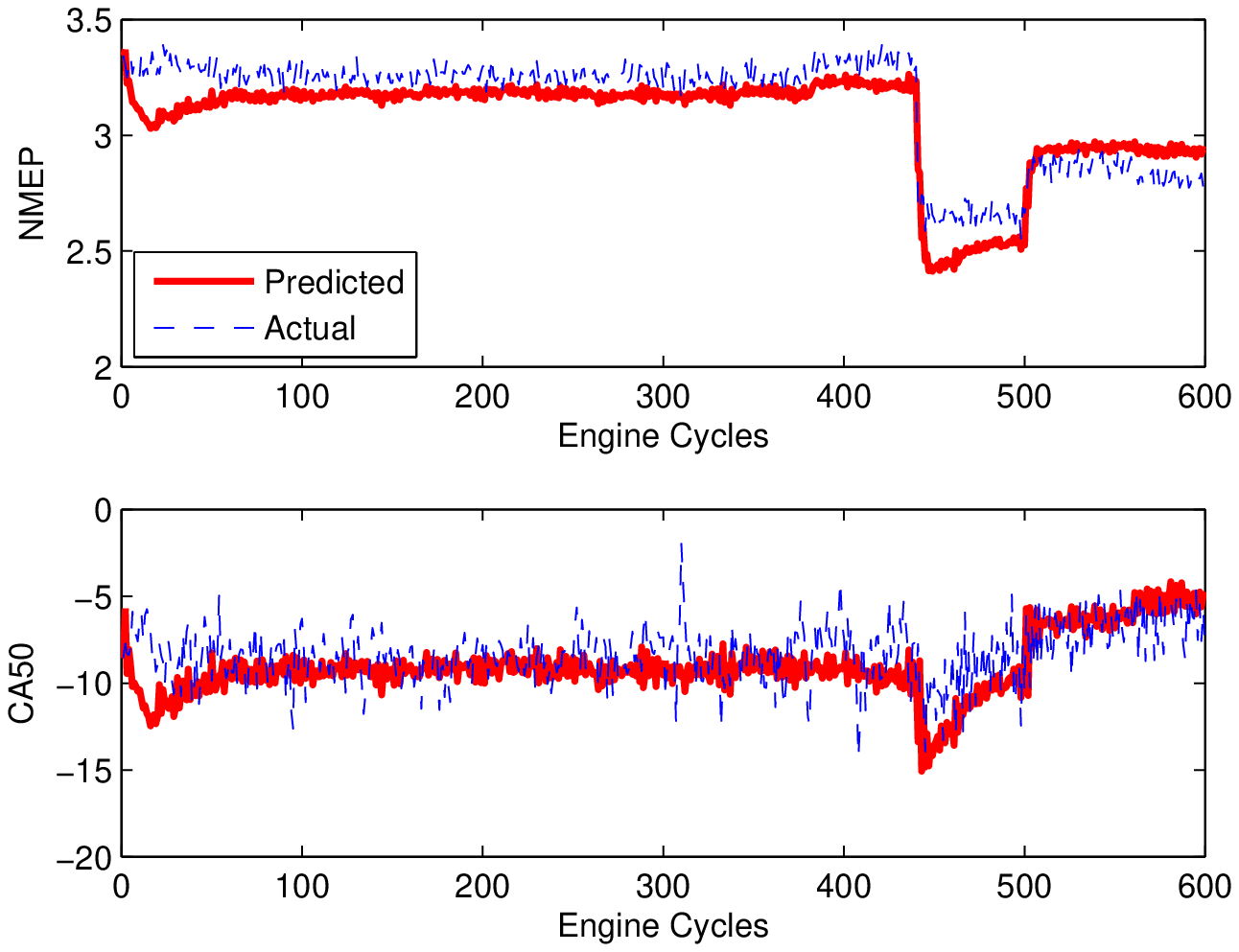}}\\
      \end{tabular}
      \caption{Prediction results of the SG-ELM algorithm showing CA50, IMEP and one input variable (fueling) for 2 unseen data sets.}
      \label{}
\end{figure*}

The MSAP predictions of the models are summarized in Figures \ref{1800_pred_1}-\ref{1800_pred_4} where model predictions for NMEP and CA50 are compared against real experimental data. Here the model is initialized using the experimental data at the first instant and allowed to make predictions recursively for several steps ahead. It can be seen that the nonlinear models outperform the linear model and at the same time the online learning models perform similar to the offline trained models indicating that online learning can fully identify the engine behavior at the operating condition where the data is collected. It should be noted that this task is a case of multi-input multi-output modeling which adds some limitations to the SG-ELM methods. When the model complexity increases, the SG-ELM require more excitations for convergence, as opposed to OS-ELM which converges more aggressively (although at the loss of stability). Further, the tuning of gradient step size $\Gamma_{SG}$ may be time-consuming for systems predicting multiple outputs with different noise characteristics. The OS-ELM on the other hand is much more elegant as there are no parameters to be tuned once properly initialized.

\section{Application Case Study 2: Online classification learning (with class imbalance) for identifying the dynamic operating envelope of an HCCI Engine}\label{cs2_sec}

The problem considered in this case study is to develop a predictive model of the dynamic operating envelope of the HCCI engine. For developing stable model based controller for HCCI engines, it is necessary to prevent the engine drifting towards instabilities such as misfire, ringing, knock, etc \cite{misf1,misf2}. To this end, a dynamic operating envelope of the HCCI engine was developed using machine learning models \cite{vijay_open}. However, the modeling was performed offline. In this paper, an online learning framework for modeling the operating envelope of HCCI engine is developed using both OS-ELM and SG-ELM algorithms.

In this paper, the operating envelope defined by two common HCCI unstable modes - a complete misfire and a high variability combustion (a more detailed description is given in section \ref{instab_sec}) is studied. The problem of identifying the HCCI operating envelope using experimental data can be posed as a classification problem. The engine sensor data can be manually labeled as stable or unstable depending on engine based heuristics. Further, the engine dynamic data consists of a large number of stable class data compared to unstable class data, which introduces an imbalance in class proportions. As a result, the problem can be posed as a class imbalance learning of a binary classification decision boundary. For class imbalance learning, a cost-sensitive approach that modifies the objective function of the learning system to weigh the minority class data more heavily, is preferred over under-sampling and over-sampling approaches \cite{vijay_open}.

Online learning algorithms using OS-ELM, SG-ELM are compared for classification performance. The above nonlinear models are compared against a baseline linear classification model and an offline trained nonlinear ELM model to make similar justifications as in the previous case study. The linear baseline model is included to justify the benefits of adopting a nonlinear model while the offline trained model is included to show the effectiveness of online algorithms in capturing the underlying behavior.

\subsection{Model Structure and Evaluation Metric}
The HCCI operating envelope is a function of the engine control inputs and engine physical variables such as temperature, pressure, flow rate etc. Also, the envelope is a dynamic system and so a predictive model requires the measurement history up to an order of $N_h$. The dynamic classifier model can be given by
\begin{equation}\label{}
\hat{y}_{k+1}=\sgn(f(x_k))
\end{equation}
where $sign(.)$ represents the sign function, $\hat{y}_{k+1}$ indicates model prediction for the future cycle $k+1$, $f(.)$ can take any structure depending on the learning algorithm and $x_k$ is given by
\begin{multline}\label{app_inp}
x_k=[IVO, EVC, FM, SOI, T_{in}, P_{in}, \dot{m}_{in}, \\
    T_{ex}, P_{ex}, T_{c}, FA, NMEP, CA50]^T
\end{multline}
at cycle $k$ upto cycle $k-N_h+1$. In the following sections, the function $f(.)$ is learned using the available engine experimental data using the two online ELM algorithms. The engine measurements and their time histories (defined by $x_k$) are considered inputs to the model while the stability labels are considered outputs. The feature vector is of dimension $n$=39 includes sensor measurements such as FM, IVO, EVC, SOI, $T_c$, $T_{in}$, $P_{in}$, $\dot{m}_{in}$, $T_{ex}$, $P_{ex}$, NMEP, CA50 and FA along with $N_h=1$ cycles of history (see \eqref{app_inp}). The engine experimental data is split into training and testing sets. The training set consists of about 14300 cycles of data processed one-by-one as sampled by the engine ECU. After the training phase, the parameter update is switched off and the models are evaluated for the next 6200 cycles of data for one step ahead classification. The ratio of number of majority class data to number minority class data ($r$) for the training set is about 4.5:1 and for the testing set is 9:1. The nonlinear model approximating $f(.)$ is initialized to an extreme learning machine model with random input layer weights and random values for the covariance matrices and output layer weights. All the nonlinear models consist of 10 hidden units with fixed randomized input layer parameters. Similar to the previous case study, a small portion of the training data is used to initialize the ELM model parameters as well as the covariance matrix. The SG-ELM parameter $\Gamma_{SG}$ is tuned to be $0.001 \quad I_{10}$ using trial and error. A weighted classification version of the algorithms is developed to handle the class imbalance problem. The minority class data is weighted higher by $r$ times $f_s$ where $r$ is the imbalance ratio of the training data and is computed online as the ratio of the number of majority class to number of minority class data until that instant.

For the class imbalance problem considered here, a conventional classifier metric like the overall misclassification rate cannot be used as it would find a biased classifier, i.e., it would find a classifier that ignores the minority class data. For instance, a data set that has 95\% of majority class data (with label +1) would achieve 95\% classification accuracy by predicting all the labels to be +1 which is obviously undesirable. Hence the following evaluation metric used for skewed data sets is considered. Let $TP$ and $TN$ represent the total number of positive and negative class data classified correctly by the classifier. If $N^+$ and $N^-$ represent the total number of positive and negative class data respectively, the true positive rate (TPR) and true negative rate (TNR), geometric mean (GM) of TPR and TNR, and the total accuracy (TA) of the classifier can be defined as follows \cite{toh_imbalanced}. It should be noted that the total accuracy and geometric mean weights the accuracy of majority and minority classes equally, i.e., they have high values only when both classes of data are classified correctly.

\begin{eqnarray}
\nonumber TPR &=& \frac{TP}{N^+} \\
\nonumber TNR &=& \frac{TN}{N^-} \\
\nonumber GM &=& \sqrt{TPR \times TNR} \\
TA &=& 0.5(TPR+TNR).
\end{eqnarray}

\subsection{Results and Discussion}
\begin{figure*}[]
      \centering
      \begin{tabular}{cc}
      \subfloat[OS-ELM (dataset 1)]{\label{}\includegraphics[scale=0.6]{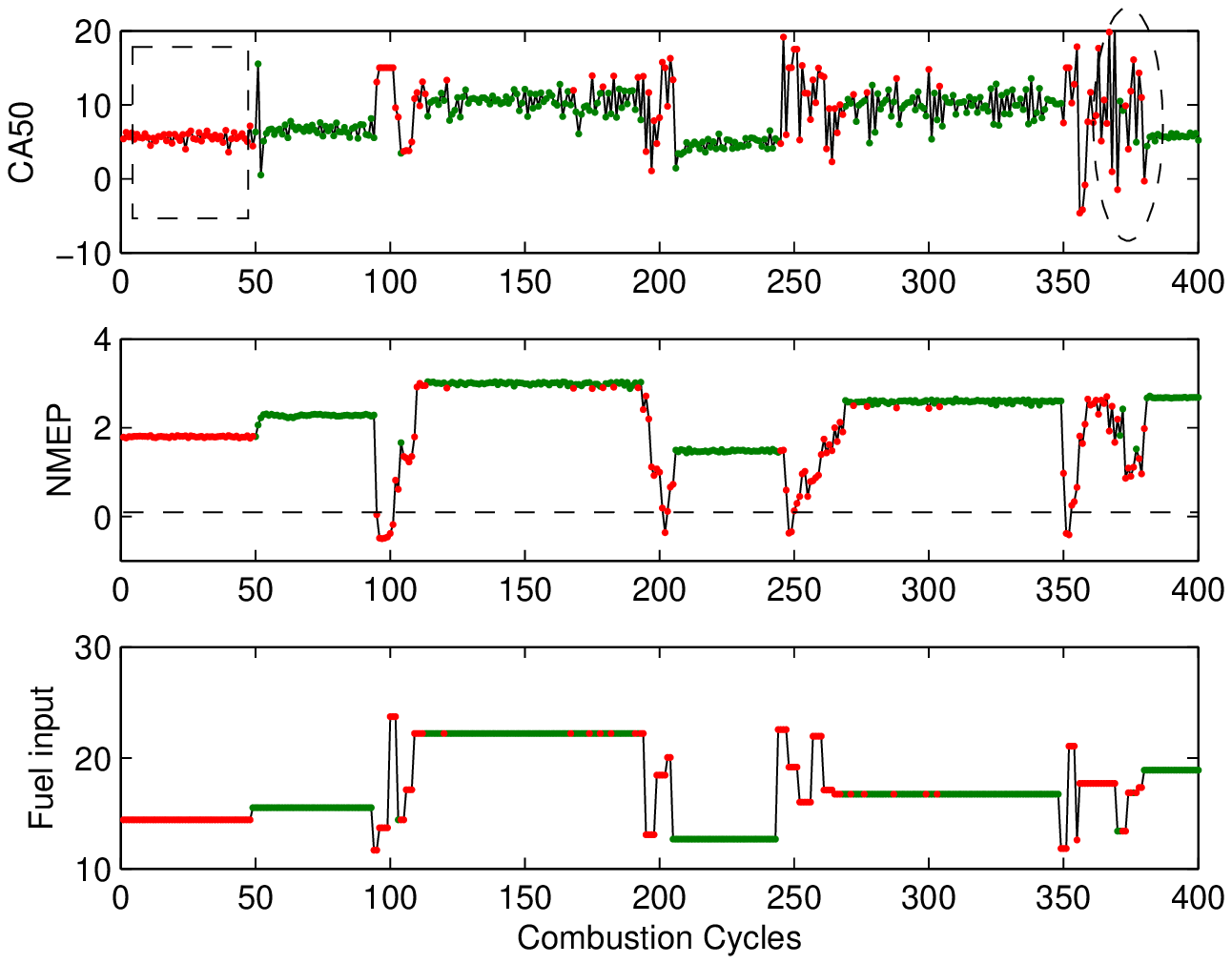}}
      &      \subfloat[OS-ELM (dataset 2)]{\label{}\includegraphics[scale=0.6]{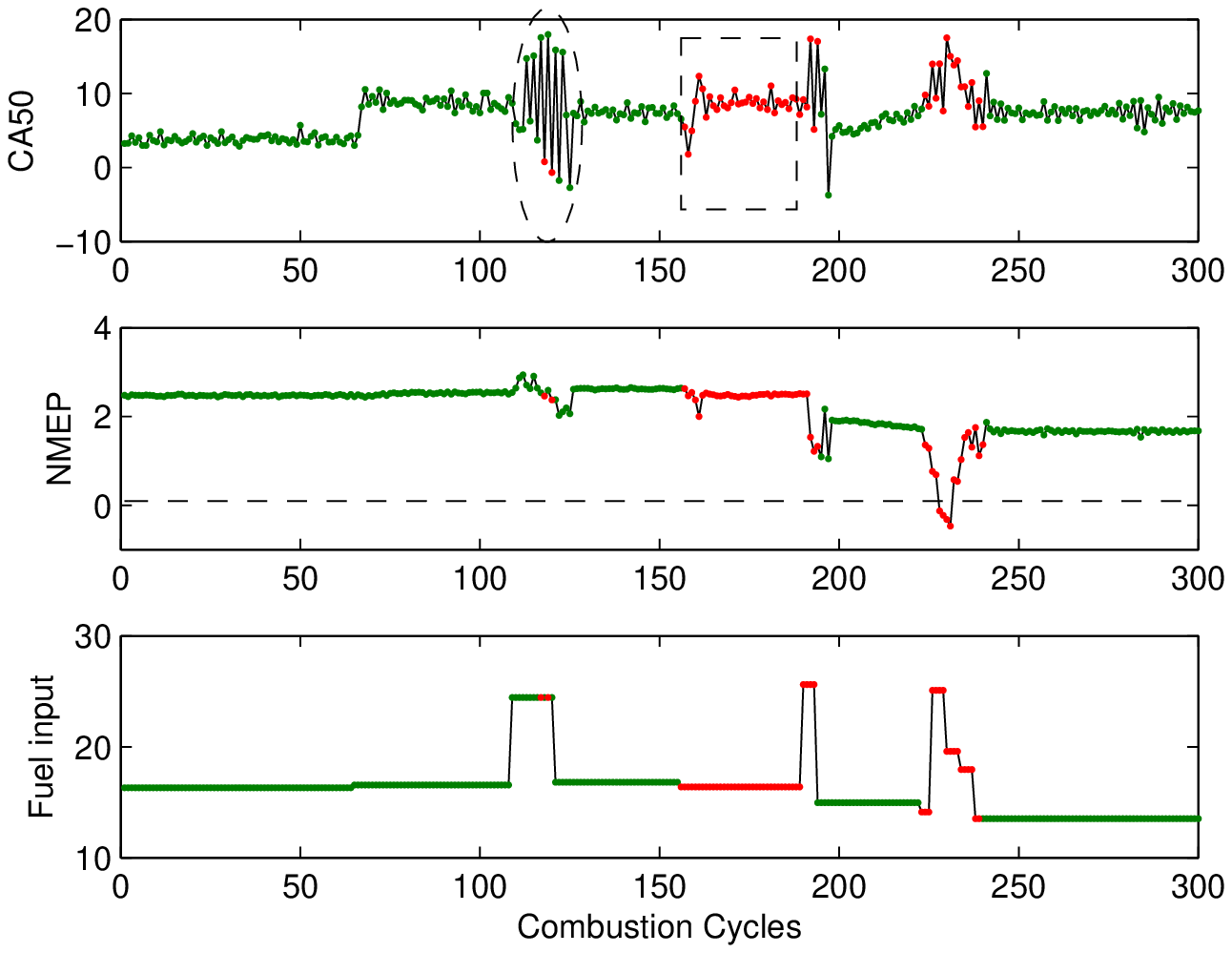}}\\
      \subfloat[SG-ELM (dataset 1)]{\label{}\includegraphics[scale=0.6]{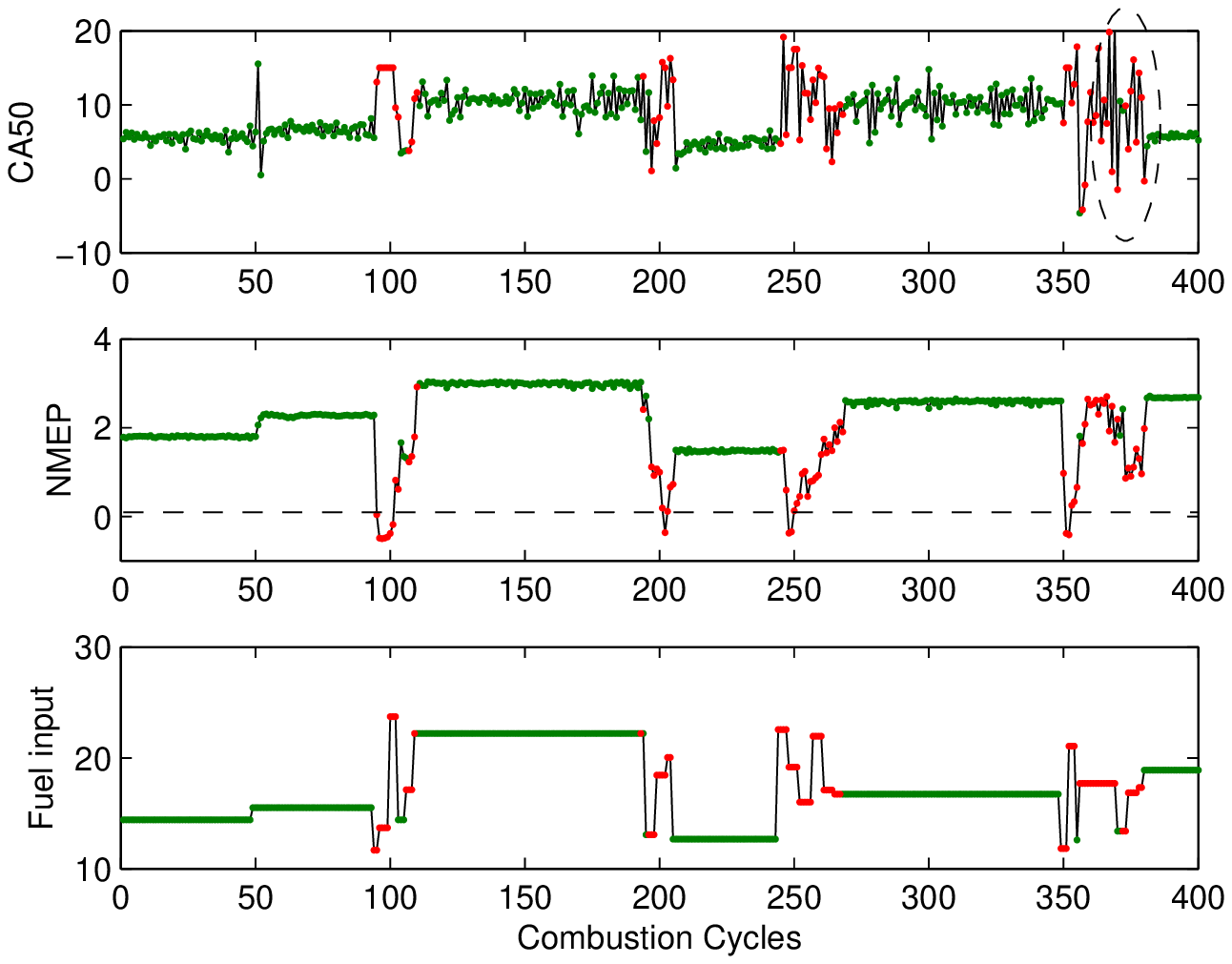}}
      &      \subfloat[SG-ELM (dataset 2)]{\label{}\includegraphics[scale=0.6]{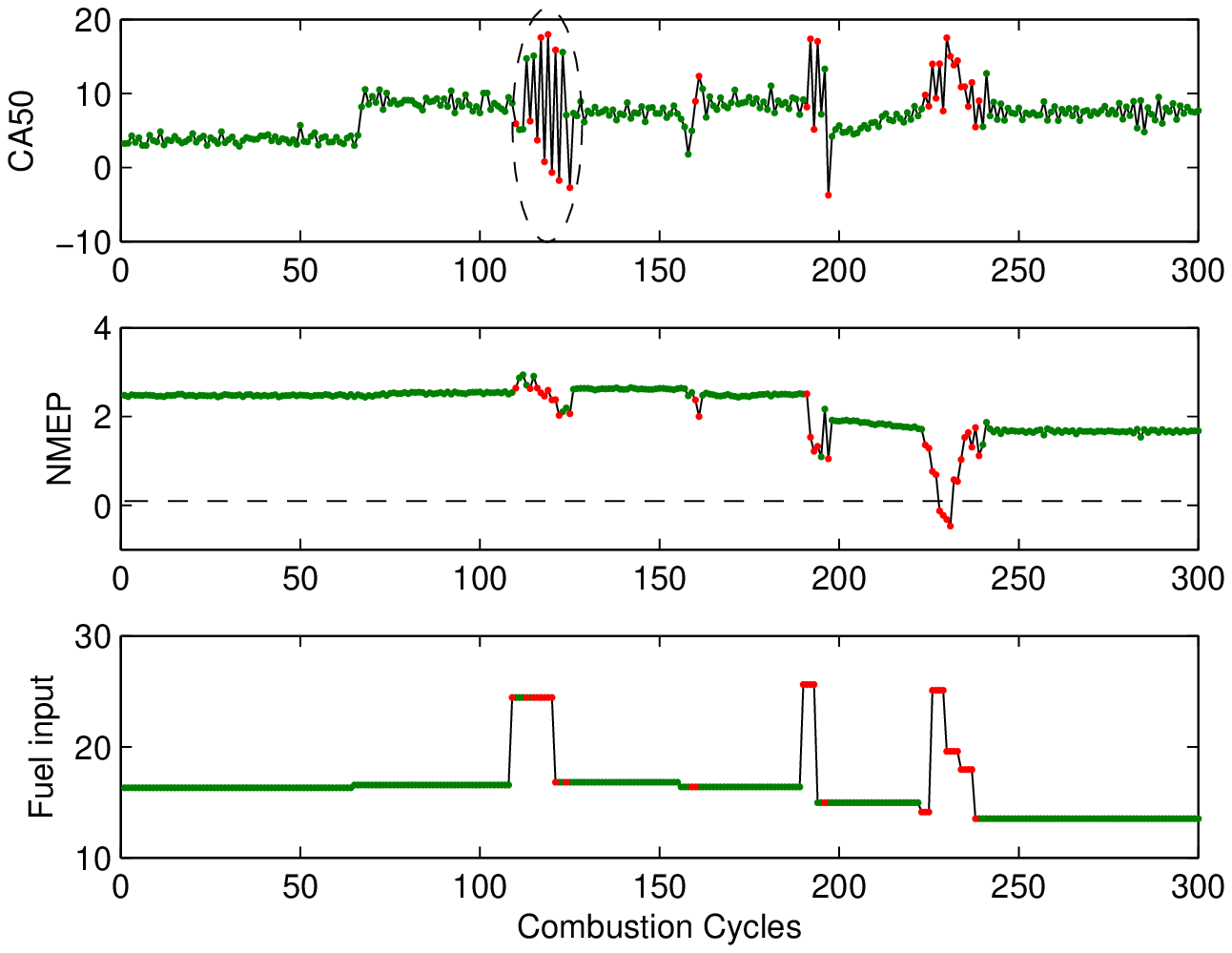}}\\
      \end{tabular}
      \caption{Classification results of OS-ELM and SG-ELM models showing CA50, IMEP and one input variable (fueling) for 2 unseen data sets. The color code indicates model prediction - green (and red) indicate stable (and unstable) prediction by the model. The dotted line in the IMEP plot indicates misfire limit, dotted ellipse in CA50 plot indicates high variability instability mode while dotted rectangle indicates a wrong predictions by model.}
      \label{engine_class_pred}
\end{figure*}

The results of online imbalance classification can be summarized in Table \ref{class_ol_tab} where computational time as well as classification performance can be compared. It can be observed that the developed classification models perform well for the HCCI boundary identification problem (see average accuracies of all models are above 80\%). The problem is mildly nonlinear as linear models achieve similar accuracies as that of their nonlinear counterparts. Both OS-ELM and SG-ELM perform well and achieve results similar to an offline model indicating completeness of learning. The SG-ELM has a slight advantage in terms of computational efficiency. The algorithm is simple and requires about half of the time required to train an OS-ELM model. Further, for the considered classification problem, the prediction accuracy of SG-ELM is slightly better than OS-ELM indicating the suitability of SGD based online learning for the HCCI problem. A subtle advantage observed for the OS-ELM is that, although the combined accuracy is slightly inferior to that of the SG-ELM, the accuracies of the positive examples and negative examples are very close to each other indicating that the model is well balanced to predict both majority class as well as minority class data well. The SG-ELM on the other hand, in spite of fine-tuning the parameters, fails to achieve this. A further tuning can be done to improve the accuracy of a particular class of data, typically sacrificing some accuracy predicting the other. The predictions of the online SG-ELM model is shown in Fig. \ref{engine_class_pred}.

\begin{table}[htbp]
  \centering
  \caption{Performance comparison of the nonlinear models (OS-ELM and SG-ELM) for the online class imbalance learning problem. A baseline linear model and an offline trained ELM model (O-ELM) are also used for comparison.}
    \begin{tabular}{cccccc}
    \hline
    Algorithms & Training & TPR   & TNR   & Total & GM \\
          & Time in s &       &       & Accuracy & Accuracy \\
    \hline
    Linear & 0.58  & 0.9982  & 0.6374  & 0.8178  & 0.7977 \\
    OS-ELM & 0.58  & 0.8328  & 0.8341  & 0.8335  & 0.8335 \\
    SG-ELM & \textbf{0.30}  & \textbf{0.9876}  & 0.7707  & \textbf{0.8792}  & \textbf{0.8725} \\
    O-ELM & -  & 0.8265  & \textbf{0.8569}  & 0.8417  & 0.8416 \\
    \hline
    \end{tabular}%
  \label{class_ol_tab}%
\end{table}%

The models developed using OS-ELM and SG-ELM algorithms are used to make predictions on unseen engine inputs and class predictions are summarized in Fig. \ref{engine_class_pred}, while quantitative results are included in Table \ref{class_ol_tab}. As mentioned earlier, the operating envelope is a decision boundary in the input space within which any input operates the HCCI in a stable manner and any input outside the envelope might operate the engine in an unstable manner. The HCCI state variables such as NMEP, CA50 and engine sensor observations such as $T_{in}, P_{in}, \dot{m}_{in}, T_{ex}, P_{ex}, T_{c}$ at time instant $k$, along with engine control inputs such as FM, EVC, SOI at time instant $k+1$, are given as input to the models (see \eqref{app_inp}). The model predictions at time $k+1$ are obtained. The engine's actual response at time $k+1$ is also recorded. A data point is marked in red if the model predicts the engine operation to be unstable (-1) while it is marked in green if the model predicts the data point to be stable (+1). In the figures, a dotted line in the NMEP plot indicates the misfire limit, a dotted ellipse in CA50 plot indicates high variability instability mode while a dotted rectangle indicates misclassified predictions by model. To understand the variation of NMEP and CA50 with changes in control inputs, the fueling input (abbreviated as FM) is also included in the plots. It should be understood that FM is not the only input for prediction and the signals are defined as in equation \eqref{app_inp} but only the fueling input is shown in the plots owing to space constraints.

It can be seen from the above plots that as a whole, both OS-ELM and SG-ELM models classify the HCCI engine data fairly well in spite of the high amplitude noise inherent in the HCCI experimental data. The data consists of step changes in FM, EVC and SOI and whenever a `bad' combination of inputs is chosen, the engine either misfires completely (see NMEP fall below misfire limit) or exhibits high variability combustion (see dotted ellipses). The goal of this work as stated previously, is to predict if a future HCCI combustion event is stable or unstable based on available measurements. The results summarized in Table \ref{class_ol_tab} indicates that the developed models indeed accomplished the goal with a reasonable accuracy. From Fig. \ref{engine_class_pred}, it is observed that the OS-ELM has some clear misclassifications in predicting stable class data (see dotted rectangles in the plots) while this is not observed for SG-ELM. This is not surprising as the true positive rate of OS-ELM model is much lesser compared to that of SG-ELM (see Table \ref{class_ol_tab}). On the other hand, the SG-ELM has an inferior accuracy in predicting the unstable modes but is not clearly evident in the data sets used in Fig. \ref{engine_class_pred} .

\section{Conclusion} \label{concl}
A stochastic gradient descent based online learning algorithm for ELM has been developed, that guarantees stability in parameter estimation suitable for control purposes. Further, the SG-ELM demands less computation compared to the OS-ELM algorithm, as the covariance estimation step is eliminated. A stability proof is developed based on Lyapunov approach. However, the SG-ELM algorithm might involve tedious tuning of step-size parameter as well as suffer from slow convergence.

The SG-ELM and OS-ELM algorithms are applied to develop models for state variables and dynamic operating envelope of a HCCI engine to assist in model based control. The results from this article suggest that good generalization performance can be achieved using both OS-ELM and SG-ELM methods but the SG-ELM might have an advantage in terms of stability, crucial for designing robust control systems.

Although the SG-ELM appears to perform well in the HCCI identification problem, a comprehensive analysis and evaluation on several benchmark data sets is required and will be considered for future. From an application perspective, interesting areas for exploration include implementing the algorithm in real-time hardware, exploring a wide operating range of HCCI operation and development of controllers.

\section*{Acknowledgment}
This material is based upon work supported by the Department of Energy and performed as a part of the ACCESS project consortium (Robert Bosch LLC, AVL Inc., Emitec Inc.) under the direction of PI Hakan Yilmaz, Robert Bosch, LLC. X. Nguyen is supported in part by NSF Grants CCF-1115769 and ACI-1047871.

\section*{Disclaimer}
This report was prepared as an account of work sponsored by an agency of the United States Government.  Neither the United States Government nor any agency thereof, nor any of their employees, makes any warranty, express or implied, or assumes any legal liability or responsibility for the accuracy, completeness, or usefulness of any information, apparatus, product, or process disclosed, or represents that its use would not infringe privately owned rights.  Reference herein to any specific commercial product, process, or service by trade name, trademark, manufacturer, or otherwise does not necessarily constitute or imply its endorsement, recommendation, or favoring by the United States Government or any agency thereof.  The views and opinions of authors expressed herein do not necessarily state or reflect those of the United States Government or any agency thereof.

\ifCLASSOPTIONcaptionsoff
  \newpage
\fi

\bibliographystyle{IEEEtran}
\bibliography{references}

\begin{thebibliography}{10}
\providecommand{\url}[1]{#1}
\csname url@samestyle\endcsname
\providecommand{\newblock}{\relax}
\providecommand{\bibinfo}[2]{#2}
\providecommand{\BIBentrySTDinterwordspacing}{\spaceskip=0pt\relax}
\providecommand{\BIBentryALTinterwordstretchfactor}{4}
\providecommand{\BIBentryALTinterwordspacing}{\spaceskip=\fontdimen2\font plus
\BIBentryALTinterwordstretchfactor\fontdimen3\font minus
  \fontdimen4\font\relax}
\providecommand{\BIBforeignlanguage}[2]{{%
\expandafter\ifx\csname l@#1\endcsname\relax
\typeout{** WARNING: IEEEtran.bst: No hyphenation pattern has been}%
\typeout{** loaded for the language `#1'. Using the pattern for}%
\typeout{** the default language instead.}%
\else
\language=\csname l@#1\endcsname
\fi
#2}}
\providecommand{\BIBdecl}{\relax}
\BIBdecl

\bibitem{oselm}
N.~Liang, G.~Huang, P.~Saratchandran, and N.~Sundararajan, ``A fast and
  accurate online sequential learning algorithm for feedforward networks,''
  \emph{Neural Networks, IEEE Transactions on}, vol.~17, no.~6, pp. 1411--1423,
  2006.

\bibitem{thring1}
R.~Thring, ``Homogeneous-charge compression-ignition engines,'' 1989, {SAE}
  paper 892068.

\bibitem{Christensen2}
M.~Christensen, P.~Einewall, and B.~Johansson, ``Homogeneous charge compression
  ignition using iso-octane, ethanol and natural gas- a comparison to spark
  ignition operation,'' in \emph{International Fuels \& Lubricants Meeting \&
  Exposition}, Tulsa, OK, USA, Oct 1997, {SAE} paper 972874.

\bibitem{Aoyama3}
T.~Aoyama, Y.~Hattori, J.~Mizuta, and Y.~Sato, ``An experimental study on
  premixed-charge compression ignition gasoline engine,'' in
  \emph{International Congress \& Exposition}, Detroit, MI, USA, Feb 1996,
  {SAE} paper 960081.

\bibitem{kyoungjoon}
K.~Chang, A.~Babajimopoulos, G.~A. Lavoie, Z.~S. Filipi, and D.~N. Assanis,
  ``Analysis of load and speed transitions in an {HCCI} engine using 1-d cycle
  simulation and thermal networks.''\hskip 1em plus 0.5em minus 0.4em\relax
  {SAE} International, 04 2006.

\bibitem{Johansson2010}
J.~Bengtsson, P.~Strandh, R.~Johansson, P.~Tunestal, and B.~Johansson, ``Model
  predictive control of homogeneous charge compression ignition {(HCCI)} engine
  dynamics,'' in \emph{2006 IEEE International Conference on Control
  Applications}, 2006.

\bibitem{heatedintake}
Y.~Wang, S.~Makkapati, M.~Jankovic, M.~Zubeck, and D.~Lee, ``Control oriented
  model and dynamometer testing for a single-cylinder, heated-air hcci
  engine.''\hskip 1em plus 0.5em minus 0.4em\relax {SAE} International, 04
  2009.

\bibitem{nonlin_HCCI}
C.-J. Chiang, A.~Stefanopoulou, and M.~Jankovic, ``Nonlinear observer-based
  control of load transitions in homogeneous charge compression ignition
  engines,'' \emph{Control Systems Technology, IEEE Transactions on}, vol.~15,
  no.~3, pp. 438--448, 2007.

\bibitem{narrow1}
Y.~Urata, M.~Awasaka, J.~Takanashi, T.~Kakinuma, T.~Hakozaki, and A.~Umemoto,
  ``A study of gasoline-fuelled {HCCI} engine equipped with an electromagnetic
  valve train.''\hskip 1em plus 0.5em minus 0.4em\relax {SAE} International, 06
  2004.

\bibitem{narrow2}
R.~Scaringe, C.~Wildman, and W.~K. Cheng, ``On the high load limit of boosted
  gasoline {HCCI} engine operating in {NVO} mode,'' \emph{{SAE} Int. J.
  Engines}, vol.~3, pp. 35--45, 04 2010.

\bibitem{chiang2010}
C.~Chiang and C.~Chen, ``Constrained control of homogeneous charge compression
  ignition {(HCCI}) engines,'' in \emph{5th IEEE Conference on Industrial
  Electronics and Applications (ICIEA)}, 2010.

\bibitem{Ravi2009}
H.~H. L. A. F. J. C. F. C. S.~P. N.~Ravi, M. J.~Roelle and J.~C. Gerdes,
  ``Model-based control of {HCCI} engines using exhaust recompression,'' in
  \emph{IEEE Transactions on Control Systems Technology}, 2009.

\bibitem{vijay_asoc}
V.~M. Janakiraman, X.~Nguyen, and D.~Assanis, ``Nonlinear identification of a
  gasoline {HCCI} engine using neural networks coupled with principal component
  analysis,'' \emph{Applied Soft Computing}, vol.~13, no.~5, pp. 2375 -- 2389,
  2013.

\bibitem{vijay_springer}
V.~M. Janakiraman, X.~Nguyen, J.~Sterniak, and D.~Assanis,
  ``\BIBforeignlanguage{English}{A system identification framework for modeling
  complex combustion dynamics using support vector machines},'' in
  \emph{\BIBforeignlanguage{English}{Informatics in Control, Automation and
  Robotics}}, ser. Lecture Notes in Electrical Engineering.\hskip 1em plus
  0.5em minus 0.4em\relax Springer International Publishing, 2014, vol. 283,
  pp. 297--313.

\bibitem{vijay_open}
V.~Janakiraman, X.~Nguyen, J.~Sterniak, and D.~Assanis, ``Identification of the
  dynamic operating envelope of {HCCI} engines using class imbalance
  learning,'' \emph{Neural Networks and Learning Systems, IEEE Transactions
  on}, vol.~PP, no.~99, pp. 1--1, 2014.

\bibitem{misf1}
G.~T. Kalghatgi and R.~A. Head, ``Combustion limits and efficiency in a
  homogeneous charge compression ignition engine,'' \emph{Int. J. Engine Res},
  vol.~7, pp. 215--236, 2006.

\bibitem{misf2}
M.~Shahbakhti and C.~R. Koch, ``Characterizing the cyclic variability of
  ignition timing in a homogenous charge compression ignition engine fueled
  with n-heptane/iso-octane blend fuels,'' \emph{Int. J. Engine Res}, vol.~9,
  pp. 361--397, 2008.

\bibitem{svm_ilya}
I.~V. Kolmanovsky and E.~G. Gilbert, ``Support vector machine-based
  determination of gasoline direct injected engine admissible operating
  envelope.''\hskip 1em plus 0.5em minus 0.4em\relax {SAE} International, 03
  2002.

\bibitem{misfire_emissions}
A.~Soliman, G.~Rizzoni, and V.~Krishnaswami, ``The effect of engine misfire on
  exhaust emission levels in spark ignition engines.''\hskip 1em plus 0.5em
  minus 0.4em\relax {SAE} International, 02 1995.

\bibitem{misfire_emission}
P.~Azzoni, D.~Moro, C.~M. Porceddu-cilione, and G.~Rizzoni, ``Misfire detection
  in a high-performance engine by the principal component analysis
  approach.''\hskip 1em plus 0.5em minus 0.4em\relax {SAE} International, 1996.

\bibitem{vijay_control}
V.~Janakiraman, X.~Nguyen, J.~Sterniak, and D.~Assanis, ``Nonlinear model
  predictive control of gasoline hcci engines using extreme learning
  machines,'' \emph{Neural Networks and Learning Systems, IEEE Transactions
  on}, vol.~PP, no.~99, pp. 1--1, In Review.

\bibitem{illcond1}
G.~Zhao, Z.~Shen, C.~Miao, and Z.~Man, ``On improving the conditioning of
  extreme learning machine: A linear case,'' in \emph{Information,
  Communications and Signal Processing, 2009. ICICS 2009. 7th International
  Conference on}, dec. 2009, pp. 1 --5.

\bibitem{illcond2}
F.~Han, H.-F. Yao, and Q.-H. Ling, ``An improved extreme learning machine based
  on particle swarm optimization,'' in \emph{Bio-Inspired Computing and
  Applications}, ser. Lecture Notes in Computer Science.

\bibitem{illcond3}
M.~T. Hoang, H.~Huynh, N.~Vo, and Y.~Won, ``A robust online sequential extreme
  learning machine,'' in \emph{Advances in Neural Networks}, ser. Lecture Notes
  in Computer Science.

\bibitem{illcond4}
H.~T. Huynh and Y.~Won, ``Regularized online sequential learning algorithm for
  single-hidden layer feedforward neural networks,'' \emph{Pattern Recognition
  Letters}, vol.~32, no.~14, pp. 1930 -- 1935, 2011.

\bibitem{pred_adap}
V.~Akpan and G.~Hassapis, ``Adaptive predictive control using recurrent neural
  network identification,'' in \emph{Control and Automation, 2009. MED '09.
  17th Mediterranean Conference on}, june 2009, pp. 61 --66.

\bibitem{lyap_rbf}
L.~Yan, N.~Sundararajan, and P.~Saratchandran, ``Nonlinear system
  identification using lyapunov based fully tuned dynamic rbf networks,''
  \emph{Neural Process. Lett.}, vol.~12, no.~3, pp. 291--303, Dec. 2000.

\bibitem{glomap}
P.~Singla and J.~Junkins, \emph{Multi-Resolution Methods for Modeling and
  Control of Dynamical Systems}, ser. Chapman \& Hall/CRC Applied Mathematics
  \& Nonlinear Science.\hskip 1em plus 0.5em minus 0.4em\relax Taylor \&
  Francis, 2010.

\bibitem{4Huang2005}
G.-B. Huang, Q.-Y. Zhu, and C.-K. Siew, ``Extreme learning machine: Theory and
  applications,'' \emph{Neurocomputing}, vol.~70, pp. 489--501, 2006.

\bibitem{huang12}
G.-B. Huang, H.~Zhou, X.~Ding, and R.~Zhang, ``Extreme learning machine for
  regression and multiclass classification.'' \emph{IEEE Transactions on
  Systems, Man, and Cybernetics, Part B}, vol.~42, no.~2, pp. 513--529, 2012.

\bibitem{sgd_bottou}
L.~Bottou, ``\BIBforeignlanguage{English}{Large-scale machine learning with
  stochastic gradient descent},'' in
  \emph{\BIBforeignlanguage{English}{Proceedings of COMPSTAT'2010}},
  Y.~Lechevallier and G.~Saporta, Eds.\hskip 1em plus 0.5em minus 0.4em\relax
  Physica-Verlag HD, 2010, pp. 177--186.

\bibitem{sgd_exprate}
N.~{Le Roux}, M.~Schmidt, and F.~Bach, ``{A Stochastic Gradient Method with an
  Exponential Convergence Rate for Strongly-Convex Optimization with Finite
  Training Sets},'' INRIA, Tech. Rep. arXiv:1202.6258v1, 2012.

\bibitem{vap95}
V.~Vapnik, \emph{The Nature of Statistical Learning Theory}.\hskip 1em plus
  0.5em minus 0.4em\relax New York: Springer, 1995.

\bibitem{Shalev-Shwartz_sgd}
S.~Shalev-Shwartz and N.~Srebro, ``Svm optimization: inverse dependence on
  training set size,'' in \emph{Proceedings of the 25th international
  conference on Machine learning}, ser. ICML '08.\hskip 1em plus 0.5em minus
  0.4em\relax New York, NY, USA: ACM, 2008, pp. 928--935.

\bibitem{jingsun}
P.~Ioannou and J.~Sun, \emph{Robust adaptive control}.

\bibitem{disc_barb_lemma}
J.~Spooner, M.~Maggiore, R.~Ord{\'o}{\~n}ez, and K.~Passino, \emph{Stable
  Adaptive Control and Estimation for Nonlinear Systems: Neural and Fuzzy
  Approximator Techniques}, ser. Adaptive and Learning Systems for Signal
  Processing, Communications and Control Series.\hskip 1em plus 0.5em minus
  0.4em\relax Wiley, 2004.

\bibitem{hcci_book}
F.~Zhao, T.~N. Asmus, D.~N. Assanis, J.~E. Dec, J.~A. Eng, and P.~M. Najt,
  \emph{Homogeneous Charge Compression Ignition (HCCI) Engines}.\hskip 1em plus
  0.5em minus 0.4em\relax {SAE} International, March 2003.

\bibitem{george}
G.~A. Lavoie, J.~Martz, M.~Wooldridge, and D.~Assanis, ``A multi-mode
  combustion diagram for spark assisted compression ignition,''
  \emph{Combustion and Flame}, vol. 157, no.~6, pp. 1106 -- 1110, 2010.

\bibitem{Nelles17}
O.~Nelles, \emph{Nonlinear System Identification: From Classical Approaches to
  Neural Networks and Fuzzy Models}.\hskip 1em plus 0.5em minus 0.4em\relax
  Springer, 2001.

\bibitem{narendra}
K.~S. Narendra and K.~Parthasarathy, ``Identification and control of dynamical
  systems using neural networks,'' vol.~1, no.~1, pp. 4--27, Mar. 1990.

\bibitem{auto_mpc}
L.~Re, F.~Allg{\"o}wer, L.~Glielmo, C.~Guardiola, and I.~Kolmanovsky,
  \emph{Automotive Model Predictive Control: Models, Methods and Applications},
  ser. Lecture Notes in Control and Information Sciences.\hskip 1em plus 0.5em
  minus 0.4em\relax Springer, 2010.

\bibitem{vapnik}
V.~Vapnik, \emph{The Nature of Statitical Learning Theory}.\hskip 1em plus
  0.5em minus 0.4em\relax Springer-Verlag GmbH, 1995.

\bibitem{toh_imbalanced}
K.-A. Toh, ``Deterministic neural classification,'' \emph{Neural Comput.},
  vol.~20, no.~6, pp. 1565--1595, Jun. 2008.

\end{thebibliography}

\end{document}